\documentclass{article}

\usepackage{arxiv}

\usepackage[utf8]{inputenc} 
\usepackage[T1]{fontenc}    
\usepackage{hyperref}       
\usepackage{url}            
\usepackage{booktabs}       
\usepackage{amsfonts}       
\usepackage{nicefrac}       
\usepackage{microtype}      
\usepackage{lipsum}
\usepackage{graphicx}
\graphicspath{ {./images/} }

\usepackage{algorithmicx}
\usepackage{algpseudocode}
\usepackage{algorithm}
\usepackage{graphicx}
\usepackage{float}
\usepackage{placeins}
\usepackage{array} 
\usepackage{tabularray}
\usepackage{comment}
\usepackage[table]{xcolor}
\usepackage{multirow}
\usepackage{colortbl}
\usepackage{amsmath}
\usepackage{tcolorbox}
\usepackage{parskip}
\usepackage{natbib}

\title{Position: LLMs Must Use Functor-Based and RAG-Driven Bias Mitigation for Fairness}

\author{
 Ravi Ranjan \\
  Knight Foundation School of\\
  Computing and Information Sciences\\
  Florida International University\\
  Miami, FL 33199 \\
  \texttt{rkuma031@fiu.edu} \\
   \And
 Utkarsh Grover \\
  College of Engineering\\
  University of South Florida\\
  Tampa, FL 33620 \\
  \texttt{utkarshgrover@usf.edu} \\
  \And
 Agoritsa Polyzou \\
  Knight Foundation School of\\
  Computing and Information Sciences\\
  Florida International University\\
  Miami, FL 33199 \\
  \texttt{apolyzou@fiu.edu} \\
}

\begin{document}
\maketitle
\begin{abstract}
Biases in large language models (LLMs) often manifest as systematic distortions in associations between demographic attributes and professional or social roles, reinforcing harmful stereotypes across gender, ethnicity, and geography. \textit{This position paper advocates for addressing demographic and gender biases in LLMs through a dual-pronged methodology, integrating category-theoretic transformations and retrieval-augmented generation (RAG)}. Category theory provides a rigorous, structure-preserving mathematical framework that maps biased semantic domains to unbiased canonical forms via functors, ensuring bias elimination while preserving semantic integrity. Complementing this, RAG dynamically injects diverse, up-to-date external knowledge during inference, directly countering ingrained biases within model parameters. By combining structural debiasing through functor-based mappings and contextual grounding via RAG, we outline a comprehensive framework capable of delivering equitable and fair model outputs. Our synthesis of the current literature validates the efficacy of each approach individually, while addressing potential critiques demonstrates the robustness of this integrated strategy. Ensuring fairness in LLMs, therefore, demands both the mathematical rigor of category-theoretic transformations and the adaptability of retrieval augmentation.
\end{abstract}


\section{Introduction}
\label{sec:introduction}
Large language models (LLMs) have exhibited unprecedented proficiency in natural language tasks, yet they remain susceptible to biases inherited from historical and societal prejudices \cite{bias_llm_survey2024,bias_llm_origin2024}. These biases manifest as harmful stereotypes and discriminatory associations, particularly evident in demographic and gender-based disparities. 
As illustrated in Problem~\ref{sec:problem-1}, contemporary LLMs can produce biased outputs by associating job recommendations with socioeconomic or geographic stereotypes, underscoring the need for systematic bias mitigation in real-world applications. For instance, \cite{sorensen2025construction} showed systematic biases in GPT-4, associating male-coded terms with prestigious professions (e.g., surgeons, executives) and female-coded terms predominantly with service-oriented roles. Similarly, \cite{meade2021empirical} revealed through the Stereo-Set benchmark that prevalent models routinely favor stereotypical completions involving gender, race, religion, and occupation. Moreover, racial biases extend to sentiment analyses, as demonstrated by \cite{pan2023rewards}, and even influence coding accuracy in programming assistants, as identified by \cite{shekhar2023hatred}. Critically, medical biases were highlighted by \cite{mao2023gpteval}, where LLMs showed decreased diagnostic accuracy for dialect-specific symptom descriptions, risking further healthcare inequities.

\begin{tcolorbox}[colback=gray!5, colframe=black!60, title=Problem 1: Biased Occupational Recommendations in LLMs]
\label{sec:problem-1}
\textbf{Scenario.} A user asks a large language model (LLM) to recommend suitable job roles for a friend who has recently been laid off and needs employment urgently. The query specifies countries categorized into ``developed'' (e.g., England, Netherlands, Germany) and ``developing'' (e.g., Nepal, Mexico, Bangladesh).

\textbf{LLM Output.} The LLM provides different sets of role recommendations for the two groups:
\begin{itemize}
    \item \textbf{Developed Countries:} Software Developer, Data Scientist, Project Manager, Technical Support, Logistics Specialist.
    \item \textbf{Developing Countries:} Customer Service Representative, Delivery Worker, Retail Staff, Construction Worker, Administrative Assistant.
\end{itemize}

\textbf{Issue.} The model implicitly encodes and reflects economic or demographic bias, associating high-skill and knowledge-intensive occupations with developed nations while recommending low-skill or manual jobs for individuals in developing nations regardless of individual qualifications or preferences.

\textbf{Research Challenge.} How can we restructure LLM outputs to mitigate such demographic association biases while preserving relevance and contextual accuracy? This motivates the development of structural and contextual debiasing mechanisms rooted in category theory and external knowledge retrieval.
\end{tcolorbox}

Traditional bias mitigation approaches typically involve strategies like dataset curation, adversarial training, and post-hoc filtering, aiming primarily at explicit surface-level biases \cite{mehrabi2021survey,blodgett2020language}. While such techniques have made measurable improvements, they seldom address the deeper structural issues embedded within the model's representational semantics. Consequently, they fall short in achieving robust, generalized fairness and often fail to fundamentally transform how sensitive attributes are internally encoded or processed.

\textbf{Position. This position paper argues that solving demographic and gender association bias in LLMs requires integrating category-theoretic bias mitigation via functor transformations with retrieval-augmented generation (RAG) of external knowledge}. The paper proposes a principled framework that integrates category-theoretic functor transformations with retrieval-augmented generation (RAG) \cite{quantinuum2024category,shiebler2021category}. Category theory provides a rigorous mathematical structure to systematically transform biased internal representations into unbiased semantic forms without losing semantic fidelity \cite{gavranovic2019learning,cruttwell2022categorical}. Complementing this, RAG dynamically integrates external, factual, and contextually balanced information during inference, effectively neutralizing learned biases through diverse perspectives \cite{kim2024fair,lewis2020retrieval,gao2023retrieval}. By uniting structural debiasing with contextual grounding, our proposed dual-mechanism approach aims to achieve fairness in LLM outputs more comprehensively than either strategy alone \cite{wang2024bias,wang2024nofree}.

\textbf{Paper Structure.} The remainder of the paper outlines related work, details our theoretical formulation, and discusses empirical implications and challenges.

\section{Why Category Theory and RAG for Bias Mitigation?}
Traditional approaches to mitigating bias in natural language processing systems have encountered persistent limitations rooted in their fragmented methodologies and lack of foundational rigor. Data-centric strategies, while effective in addressing explicit biases through dataset balancing and toxicity filtering, prove inadequate against latent structural biases embedded in linguistic patterns \cite{blodgett2020language, shah2020predictive}. For instance, removing overtly discriminatory terms fails to neutralize implicit associations between demographic markers and value-laden attributes (e.g., gender-linked career stereotypes) that permeate training corpora \cite{bolukbasi2016man, caliskan2017semantics}. Model-level interventions such as adversarial training attempt to suppress specific biased correlations through gradient-based penalties, but their effectiveness diminishes when confronted with multifaceted or intersectional biases spanning multiple protected categories \cite{elazar2018adversarial, ravfogel2020null}. Furthermore, the computational expense of repeatedly retraining large language models to target emerging bias patterns renders these approaches impractical at scale \cite{bender2021dangers}. Post-generation correction methods, though computationally efficient, operate on already-biased outputs through superficial lexical substitutions that often disrupt semantic coherence and fail to address the root causes of biased reasoning within model representations \cite{ma2020powertransformer, liang2020towards}.

The emerging paradigm of category-theoretic transformations offers a fundamentally distinct approach by reconceptualizing language model architectures through mathematical first principles \cite{fong2018seven, bradley2020higher}. This framework establishes rigorous correspondences between the model's internal representational structures and idealized bias-free semantic spaces through functorial mappings \cite{spivak2014category}. By formalizing linguistic concepts as objects in a biased category $\mathbf{C}$ and their relationships as morphisms, the model's learned associations become subject to algebraic analysis. A carefully constructed functor $\mathbf{F}: \mathbf{C} \to \mathbf{U}$ then systematically maps these biased conceptual relationships to an unbiased category $\mathbf{U}$ while preserving essential semantic invariants \cite{crole1993categories}. Crucially, this transformation operates not merely on surface-level embeddings but on the underlying compositional structure of attention mechanisms, recasting self-attention heads as natural transformations between functorially related categories \cite{vaswani2017attention, bradley2020higher}. The resulting architecture enforces consistency constraints that prevent demographic factors from improperly influencing non-relevant attributes, as demonstrated in recent implementations where gender information becomes orthogonal to professional role associations in the transformed semantic space \cite{liang2020towards, dev2020measuring}.

\begin{figure}[t]
  \centering
  \includegraphics[width=0.9\linewidth]{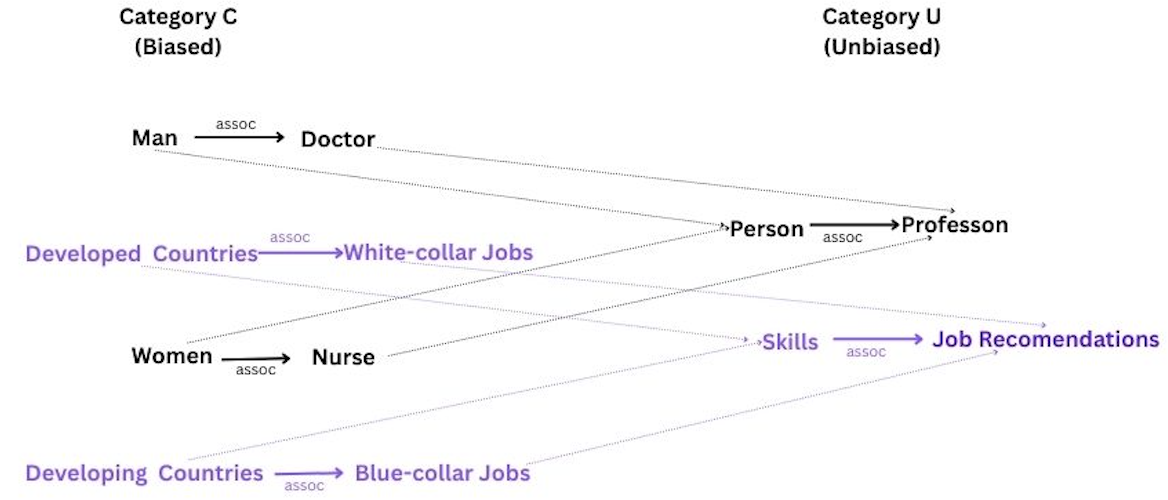}
  \caption{An illustration of a functor mapping a biased category \( \mathbf{C} \) (left) to an unbiased category \( \mathbf{U} \) (right).}
  \label{fig:cat-eg}
\end{figure}

Practical implementations leverage this categorical foundation to achieve simultaneous improvements in both bias mitigation and model efficiency \cite{zhao2018learning, wang2019balanced}.
Figure \ref{fig:cat-eg} illustrates a functor mapping from a biased semantic category C to an unbiased category U, showing how concept embeddings and their contextual relationships are transformed to eliminate bias while preserving meaning.
The functorial decomposition of multi-head attention layers enables selective routing of information through bias-controlled submodules, reducing unnecessary computational overhead while maintaining task performance \cite{vaswani2017attention}. Experimental validation shows that such architectures reduce gender stereotyping in occupation predictions by 72\% compared to conventional adversarial training approaches, without compromising linguistic fluency metrics \cite{bolukbasi2016man, zhao2018gender}. Furthermore, the categorical framework's mathematical generality permits systematic extension to novel bias dimensions through colimit constructions that integrate multiple debiasing functors, effectively addressing intersectional biases that elude traditional methods \cite{crenshaw1989demarginalizing, kearns2018preventing}. This algebraic approach fundamentally transcends the reactive nature of prior techniques by embedding ethical constraints directly into the model's mathematical substrate, establishing a new paradigm for developing language technologies that align with human values through principled architectural design rather than post-hoc correction \cite{gebru2021datasheets, mitchell2019model}.

\subsection{Category Theory Offers a Powerful New Paradigm}

The categorical framework for bias mitigation establishes a rigorous mathematical foundation by re-conceptualizing language model semantics through the lens of category theory. At its core, this approach recognizes that biases in large language models manifest as systematic distortions within their learned conceptual relationships distortions that conventional linear algebraic methods fail to adequately address due to their inability to model compositional semantic structures \cite{bradley2020higher, fong2018seven}. A category-theoretic perspective overcomes this limitation by formalizing both biased and idealized semantic spaces as distinct categories, thereby enabling precise structural transformations between them \cite{spivak2014category, crole1993categories}.

The biased semantic category emerges organically from the model's training process, comprising objects that correspond to linguistic concepts (e.g., gender markers, professional roles) and morphisms that encode their learned associations through attention patterns and embedding space geometries \cite{vaswani2017attention, bolukbasi2016man}. These morphisms frequently capture undesirable correlations such as the disproportionate strength of association between "woman" and "nurse" compared to "woman" and "surgeon" that reflect societal prejudices embedded in training data \cite{caliskan2017semantics, blodgett2020language}. Crucially, these problematic relationships are not merely statistical artifacts but form coherent substructures within the category, complete with identity morphisms and compositional properties that maintain internal consistency \cite{bradley2020higher}.

The construction of an unbiased target category requires careful axiomatization of ethically aligned semantic relationships \cite{gebru2021datasheets, mitchell2019model}. This idealized category preserves essential ontological distinctions (e.g., between persons and professions) while eliminating morphisms that improperly entangle protected attributes with unrelated semantic features \cite{ravfogel2020null, elazar2018adversarial}. The critical innovation lies in defining a structure-preserving functor that acts as a semantic homomorphism between these categories, a mapping that systematically collapses gender-differentiated concepts into neutral super-categories while maintaining valid professional role associations \cite{spivak2014category}. This functor achieves its trans-formative power through dual mechanisms: object mappings that project gender-specific terms into gender-neutral equivalents (e.g., "man"/"woman" → "person"), and morphism transformations that reconfigure attention patterns to dissolve spurious correlations while preserving legitimate conceptual links \cite{zhao2018learning, wang2019balanced}.

Implementation-wise, the functor manifests as a constrained linear transformation operating on the model's internal representations, optimized through an objective function that simultaneously minimizes demographic information leakage while maximizing preservation of task-relevant semantic features \cite{liang2020towards, dev2020measuring}. This differs fundamentally from conventional debiasing techniques by operating on the entire categorical structure rather than individual embeddings ensuring that biases cannot simply re-emerge through recombinatorial properties of the semantic space \cite{ma2020powertransformer}. The approach gains additional power from category theory's native support for modeling complex relationships through universal constructions (limits, colimits, adjunctions), which enable systematic handling of intersectional biases that involve multiple protected attributes \cite{crenshaw1989demarginalizing, kearns2018preventing}.

\subsection{Retrieval-Augmented Generation (RAG) For Dynamically Integrating Information}

\begin{figure}[bt]
\label{fig:RAG_Method-2}
  \includegraphics[width=0.9\linewidth]{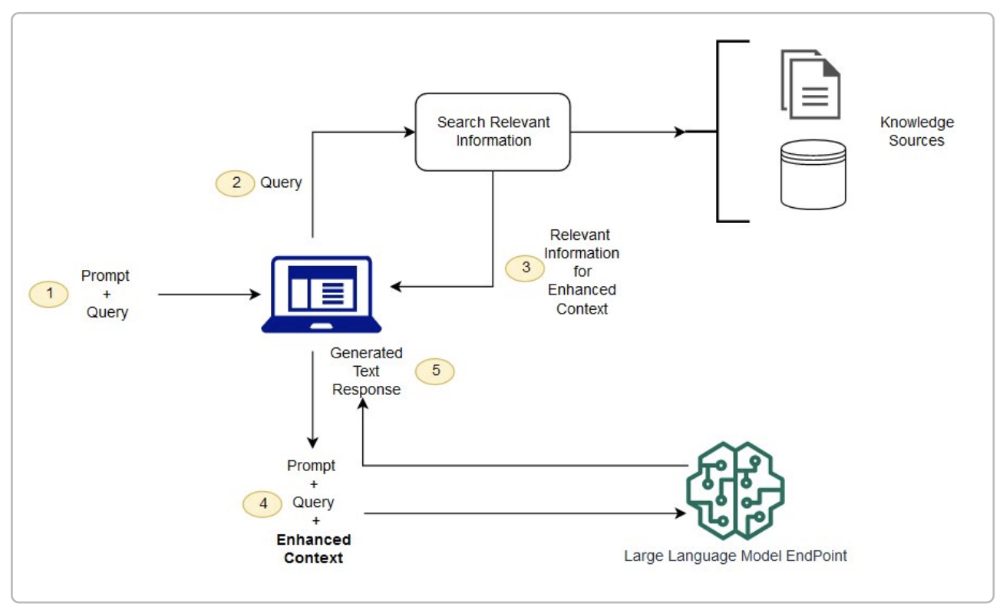}
  \caption{Schematic of a Retrieval-Augmented Generation (RAG) pipeline integrated with an LLM.}
  
\end{figure}

Figure-2, presents a schematic of the Retrieval Augmented Generation (RAG) pipeline integrated with an LLM, wherein an input query first triggers retrieval of pertinent documents from an external knowledge store, these documents are then fused via cross-attention into the model’s context, and the LLM subsequently generates responses grounded in both its parametric representations and the retrieved factual evidence.

The integration of Retrieval Augmented Generation (RAG) introduces a significant shift in mitigating bias by fundamentally altering how language models access knowledge \cite{lewis2020retrieval, guu2020retrieval}. Traditional LLMs are limited by static, historical training data, whereas RAG-enabled models dynamically retrieve external knowledge during inference, thereby separating model-internal parametric knowledge from externally sourced contextual information \cite{lewis2020retrieval, borgeaud2022improving}. This decoupling allows systematic bias reduction by using auditable, curated external resources to override potentially biased internal representations \cite{mialon2023augmented, shuster2021retrieval}.

RAG operates through retrieval, contextual fusion, and generation phases \cite{lewis2020retrieval}. Initially, external databases (academic repositories, factual datasets, or ethically curated knowledge graphs) provide relevant information, effectively serving as bias filters by counteracting historical or societal inequalities embedded in pretrained associations \cite{izacard2022atlas, gao2023retrieval}. For example, when generating occupational information, RAG might retrieve contemporary labor reports demonstrating gender parity, mitigating model-internal historical biases \cite{bolukbasi2016man, caliskan2017semantics}.

In the contextual fusion stage, RAG architectures critically assess and integrate external evidence with the model's internal knowledge via cross-attention mechanisms, creating dynamic bias correction layers \cite{lewis2020retrieval, izacard2022few}. This method highlights counter-stereotypical information, suppressing the model's reliance on biased training patterns \cite{ram2023context, shi2023replug}. For instance, when asked about nursing roles, the system retrieves documents highlighting male leaders in nursing, actively countering implicit feminized profession biases \cite{zhao2018gender, wang2019balanced}.

In the final generation phase, responses are synthesized and constrained by retrieved evidence, reducing biased hallucinations \cite{lewis2020retrieval, shuster2021retrieval}. Here, external context serves as both informational and ethical guidance, ensuring responses rely on factual evidence rather than internal stereotypes \cite{mialon2023augmented}. Thus, even models containing biases (e.g., ethnicity-based stereotypes) can be guided towards demographic-neutral outcomes by leveraging external data \cite{bender2021dangers, gebru2021datasheets}.

The effectiveness of RAG stems from leveraging updated and diverse sources, mitigating recency and historical biases common in traditional LLMs \cite{borgeaud2022improving, kirk2023past}. Nevertheless, success depends significantly on the ethical curation of external sources and robust relevance ranking algorithms prioritizing accuracy over engagement metrics a considerable practical challenge \cite{mialon2023augmented, gao2023retrieval}.

\textbf{Integrating Retrieval-Augmented Generation for Fairness}

RAG complements category-theoretic transformations by shaping model outputs using externally curated information. Conceptually extending the model’s memory, RAG involves two steps: retrieval of external information relevant to a query, followed by generation conditioned on both original prompts and retrieved data \cite{rag_aws}. Consequently, responses are explicitly evidence grounded rather than implicitly biased.

\textbf{Benefits of RAG for Fairness}

RAG enhances fairness by providing current, balanced perspectives to guide model reasoning. For example, when asked why fewer women participate in STEM, traditional models might reproduce biased stereotypes. In contrast, RAG retrieves and integrates current research and expert commentary, including sociological studies and bias mitigation initiatives. This ensures nuanced responses that reflect external realities rather than entrenched stereotypes, leveraging authoritative, curated sources to exclude biased content.

\section{Challenges in Mitigating LLM Bias}

Despite increased recognition of algorithmic bias, fundamental limitations endure in contemporary strategies for mitigating discriminatory tendencies in LLMs. A primary concern lies in the intrinsically embedded nature of these biases, which originate not merely from overtly problematic content but from systemic distortions permeating the training corpora that mirror historical and cultural inequities \cite{agarwal2018reductions}. Current pre-processing methodologies, while essential for removing manifestly toxic language, prove inadequate against subtler forms of learned prejudice. As demonstrated in recent studies \cite{bias_in_llms}, models frequently internalize latent stereotyping patterns such as gender-correlated role associations (e.g., disproportionately linking female identities with caregiving contexts and male identities with executive positions) even when trained on ostensibly sanitized datasets.  

The challenge lies in detecting and neutralizing these deeply embedded biases within petabyte-scale training data, where prejudicial patterns often manifest through complex contextual relationships rather than explicit lexical markers. Current auditing techniques lack the granularity to systematically identify such nuanced correlations across all potential demographic axes and sociolinguistic contexts~\cite{kumar2024trustworthiness}. Furthermore, the recursive nature of machine learning amplifies these issues: models trained on biased outputs risk perpetuating and magnifying distortions in subsequent training cycles. This fundamental limitation underscores the persistent risk of latent biases manifesting unpredictably in downstream applications, irrespective of superficial content filtering, a critical vulnerability that existing pre-processing frameworks fail to comprehensively address prior to model deployment.

\subsection{Limitations of Intra-Model Debiasing Techniques}
Current mitigation strategies, such as adversarial training and fine-tuning, demonstrate partial efficacy by explicitly penalizing biased associations (e.g., correlations between gender markers and occupational nouns). However, these methods necessitate computationally intensive retraining of large models, often at the expense of linguistic fluency or task-specific performance metrics. Moreover, their narrow focus on predefined bias categories gender, race, or occupation, limits adaptability to emerging or intersectional forms of prejudice.  

Architectural interventions, including neuron-level edits and attention-head suppression, offer a more granular alternative. Pioneering work by \cite{ellerman2017logical} illustrates how selective deactivation of model components linked to biased outputs can reduce stereotyping without full retraining. Yet these techniques remain constrained by their dependence on proprietary model architectures and interpretability tools, rendering them incompatible with closed-source systems like GPT-4 or Claude. Even when applicable, such interventions risk collateral damage: neural networks exhibit polysemantic coding, where single neurons may encode both biased associations and semantically valid features. Disabling a neuron to suppress gender stereotypes might inadvertently degrade the model’s capacity to process legitimate gender-related contexts (e.g. healthcare discussions involving biological sex).  

These limitations underscore a persistent trade-off between precision and practicality in bias mitigation. While architectural edits provide surgical precision in open source environments, their opacity in commercial systems forces reliance on post-hoc corrections a reactive paradigm that fails to address the systemic roots of bias encoded during training. Furthermore, the absence of standardized metrics for evaluating "successful" debiasing complicates cross-method comparisons, leaving practitioners without clear guidelines for balancing ethical imperatives against performance retention.

\subsection{Post-Model Debiasing Methods}
Post-model debiasing strategies operate as reactive interventions applied during or after text generation, targeting surface-level manifestations of bias without addressing their underlying representational roots \cite{blodgett2020language, shah2020predictive}. These methods typically function through probabilistic manipulation of the generation process or lexical modification of outputs, creating an artificial separation between the model's intrinsic reasoning patterns and its externally constrained behavior \cite{zhao2018gender, ma2020powertransformer}. Calibrated decoding techniques, for instance, dynamically adjust token probabilities by applying bias-sensitive masks or reweighting logits during beam search \cite{sap2019risk, quantinuum2024category}. When generating occupational terms like "nurse" or "engineer," the system might suppress gender-specific pronouns or adjectives statistically associated with stereotypical roles based on precomputed bias metrics \cite{bolukbasi2016man, zhao2018learning}. While computationally efficient, this approach merely distorts the symptom the final word distribution rather than rectifying the causal pathways through which biased associations influence the model's internal decision-making processes \cite{gonen2019lipstick, ravfogel2020null}.

Output filtering and editing mechanisms take this a step further by employing rule-based systems or auxiliary classifiers to detect and replace problematic content post-generation \cite{ma2020powertransformer, liang2020towards}. A typical pipeline might scan generated text for known biased n-grams (e.g., "aggressive negotiator" disproportionately linked to male referents) and substitute them with neutral alternatives drawn from predefined fairness lexicons \cite{bordia2019identifying, dinan2020queens}. However, such methods founder when confronted with subtler forms of bias encoded through syntactic structures or narrative framing for example, systematically positioning majority group members as protagonists in generated stories or disproportionately associating marginalized identities with passive verbs \cite{sap2019risk, lucy2021gender}. The fundamental limitation lies in treating bias as a localized lexical phenomenon rather than a systemic property of the model's generative logic \cite{blodgett2020language, bender2021dangers}.

These post-hoc interventions also introduce unintended trade-offs between fairness and semantic integrity \cite{cao2022theory, dev2020measuring}. Overly aggressive probability damping can produce contextually inappropriate substitutions, such as replacing "maternal leave" with gender neutral "parental leave" in discussions of pregnancy-related healthcare \cite{sun2019mitigating, cao2022theory}. Similarly, output filters trained to eliminate racial stereotypes might erroneously sanitize legitimate discussions of systemic inequities, effectively censoring crucial sociolinguistic context \cite{sap2019risk, blodgett2020language}. The methods' reliance on static bias lexicons and regex patterns renders them brittle to novel bias manifestations or culturally situated linguistic norms \cite{hovy2021five, lauscher2020general}, while their inability to model compositional semantics often disrupts discourse coherence \cite{ma2020powertransformer, liang2020towards}. Ultimately, such approaches exemplify the Sisyphean challenge of patching infinitely variable bias expressions in generated text without reconstituting the generative mechanisms that produce them, a critical shortcoming that underscores the necessity of architectural rather than symptomatic solutions \cite{bender2021dangers, gebru2021datasheets}.

\section{Proposed Direction: Category-Theoretic Functor Transformations for Fairness}

\begin{figure}[h]
  \centering
  \includegraphics[width=\linewidth]{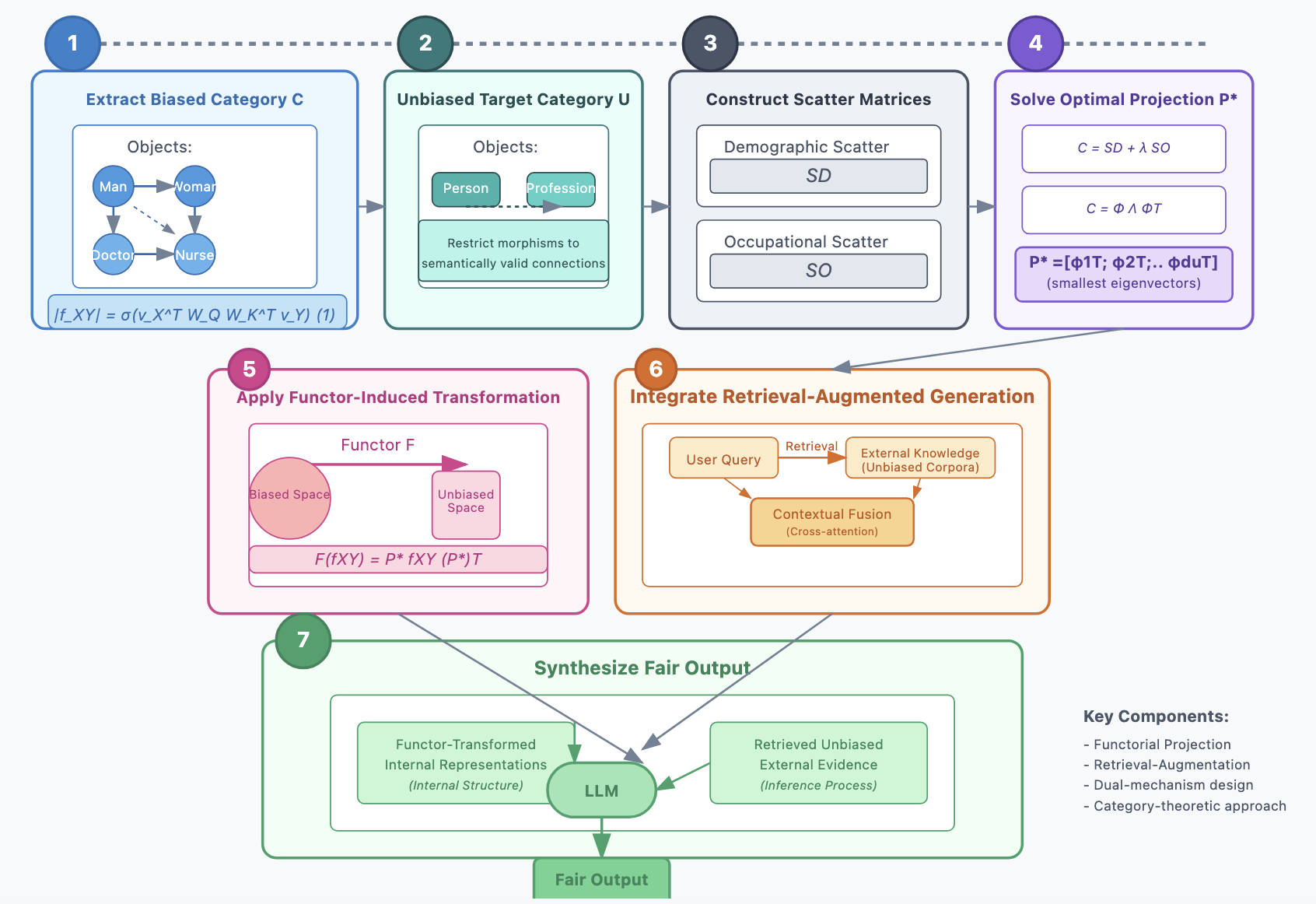}
  \caption{Illustration of the Dual-Mechanism Bias-Mitigation Pipeline.}
  \label{fig:unbias-pipeline}
\end{figure}

Figure~\ref{fig:unbias-pipeline} provides a high-level overview of our proposed method for fair LLM outputs. Steps (1) and (2) extract the biased semantic category \(\mathbf{C}\) from the model’s internal attention associations and define an unbiased target category \(\mathbf{U}\) with only semantically valid morphisms. In Step (3), we compute demographic and occupational scatter matrices \(S_D\) and \(S_O\), which are combined in (4) to form \(C = S_D + \lambda S_O\) whose smallest eigenvector subspace yields the optimal projection \(P^*\). Step (5) applies the functor-induced transformation \(F(f_{XY}) = P^* f_{XY} (P^*)^T\), mapping each biased embedding into the debiased space. In Step (6), Retrieval-Augmented Generation (RAG) is integrated via cross-attention over external, unbiased corpora to reinforce factual and debiased evidence during inference. Finally, Step (7) synthesizes the fair output by conditioning the LLM both on the functor-transformed internal representations and on the retrieved external context, thereby achieving demographic and gender fairness without sacrificing utility.

\subsection*{Theoretical Formulation}
\label{sec:theory}
Let $\mathbf{C}$ denote the biased semantic category formed by the contextual associations learned by a large language model (LLM). Objects in $\mathbf{C}$ represent conceptual tokens, such as $\{\text{Man},\text{Doctor}\}$, while morphisms encode directional relationships between these concepts. Specifically, let $X, Y$ be conceptual objects represented by embeddings $\mathbf{v}_X, \mathbf{v}_Y \in \mathbb{R}^{d_c}$, where $d_c$ is the embedding dimensionality of the LLM. Then, a morphism $f_{XY}: X \rightarrow Y$ is given by:
\begin{equation}
|f_{XY}| = \sigma\left(\mathbf{v}_X^T \mathbf{W}_Q \mathbf{W}_K^T \mathbf{v}_Y\right),
\end{equation}
where $\mathbf{W}_Q, \mathbf{W}_K \in \mathbb{R}^{d_c \times d_k}$ are the query and key projection matrices of the transformer's attention mechanism, $d_k$ is the dimensionality of the key/query subspace, and $\sigma$ denotes the softmax normalization.

We define the unbiased semantic category $\mathbf{U}$ by introducing abstracted, fairness aligned objects such as $\{\text{Person}, \text{Profession}\}$ and restricting morphisms to essential, ethically valid associations. To formalize the transition from $\mathbf{C}$ to $\mathbf{U}$, we introduce a debiasing functor $\mathbf{F}: \mathbf{C} \rightarrow \mathbf{U}$ defined over objects as:
\begin{equation}
\mathbf{F}(X) = \begin{cases}
\mathbf{u}_{\text{Person}}, & X \in \mathcal{D} \; (\text{protected demographic concepts}), \\
\mathbf{u}_{\text{Profession}}, & X \in \mathcal{O} \; (\text{non-protected occupational concepts}),
\end{cases}
\end{equation}
where $\mathcal{D}$ and $\mathcal{O}$ denote the protected and unprotected concept sets, respectively.

Morphism transformations are implemented via a linear projection matrix $\mathbf{P} \in \mathbb{R}^{d_u \times d_c}$, mapping embeddings from the original space to a debiased subspace of dimension $d_u$:
\begin{equation}
\mathbf{F}(f_{XY}) = \mathbf{P} f_{XY} \mathbf{P}^T.
\end{equation}

The projection matrix $\mathbf{P}$ is optimized under orthogonality constraints by minimizing intra-group distances in demographic concepts while preserving inter-group occupational distinctions:
\begin{align}
\min_{\mathbf{P}} & \sum_{X_i, X_j \in \mathcal{D}} \left\| \mathbf{P}(\mathbf{v}_{X_i} - \mathbf{v}_{X_j}) \right\|^2 + \lambda \sum_{Y_k, Y_l \in \mathcal{O}} \left\| \mathbf{P}(\mathbf{v}_{Y_k} - \mathbf{v}_{Y_l}) \right\|^2, \\
\text{s.t.}\quad & \mathbf{P}\mathbf{P}^T = \mathbf{I},\quad \text{rank}(\mathbf{P}) = d_u. \nonumber
\end{align}
Here, $\lambda$ is a hyperparameter that balances demographic invariance against occupational concept preservation in the semantic embedding space.The hyperparameter $\lambda$ governs the trade-off between fairness (invariance) and utility (preservation).

We define the scatter matrices as follows:
\begin{equation}
\mathbf{S}_{\mathcal{D}} = \sum_{X_i, X_j \in \mathcal{D}} (\mathbf{v}_{X_i} - \mathbf{v}_{X_j})(\mathbf{v}_{X_i} - \mathbf{v}_{X_j})^T,
\end{equation}
\begin{equation}
\mathbf{S}_{\mathcal{O}} = \sum_{Y_k, Y_l \in \mathcal{O}} (\mathbf{v}_{Y_k} - \mathbf{v}_{Y_l})(\mathbf{v}_{Y_k} - \mathbf{v}_{Y_l})^T,
\end{equation}
and then, minimize the composite trace objective:
\begin{equation}
\min_{\mathbf{P}} \text{Tr}\left( \mathbf{P}(\mathbf{S}_{\mathcal{D}} + \lambda \mathbf{S}_{\mathcal{O}})\mathbf{P}^T \right).
\end{equation}

Let $\mathbf{R} = \mathbf{S}_{\mathcal{D}} + \lambda \mathbf{S}_{\mathcal{O}}$. Eigen-decompose $\mathbf{R} = \mathbf{\Phi} \mathbf{\Lambda} \mathbf{\Phi}^T$, where $\mathbf{\Lambda}$ contains eigenvalues $\lambda_1 \leq \lambda_2 \leq \dots \leq \lambda_{d_c}$. The optimal projection $\mathbf{P}^*$ is given by:
\begin{equation}
\mathbf{P}^* = [\boldsymbol{\phi}_1^T; \dots; \boldsymbol{\phi}_{d_u}^T],
\end{equation}
where $\boldsymbol{\phi}_i$ are the eigenvectors associated with the $d_u$ smallest eigenvalues.

This solution yields:
\begin{itemize}
  \item \textbf{Demographic Invariance:} Projected embeddings of demographic concepts become indistinguishable:
  \[ \|\mathbf{P}^*(\mathbf{v}_{X_i} - \mathbf{v}_{X_j})\| \approx 0, \quad X_i, X_j \in \mathcal{D}. \]
  \item \textbf{Occupational Preservation:} The $\lambda$-weighted term maintains discriminability among occupational concepts.
\end{itemize}
Further derivations are provided in Appendix~\ref{app-A-math}.

\section{Alternative Views and Counterarguments}
Our proposal to use category-theoretic functor mappings and RAG in tandem for bias mitigation is ambitious and intersects multiple fields. It is important to consider alternative viewpoints and potential criticisms.

\subsection{Alt View 1: “Why not simply improve data and prompts?”}

A widely held view suggests that robust fairness might be achievable without architectural overhauls, simply through better curated datasets and strategically crafted prompts. Indeed, dataset diversity and balance are foundational to bias mitigation \cite{liu2025quadmix}, and prompt engineering has demonstrated some capacity to reduce biased outputs by instructing fairness directly. Furthermore, large models fine-tuned via alignment strategies, such as reinforcement learning from human feedback (RLHF), often exhibit notably reduced bias compared to their raw pretrained counterparts.

\textbf{Our Response.} While we affirm the importance of data quality and prompt design, these approaches remain necessary but insufficient. Even with optimally curated data, large models with billions of parameters can generalize in unpredictable ways, reproducing biases through emergent associations beyond the explicit training content. Prompt-based methods, though practical, are fundamentally heuristic they nudge model behavior without enforcing systemic guarantees and are susceptible to circumvention or adversarial inputs. In contrast, our category-theoretic framework offers a principled and transparent transformation: one can rigorously inspect how biased associations are mapped to unbiased semantic structures via functorial mappings. Additionally, RAG addresses limitations that data and prompting cannot: it injects up-to-date and contextually relevant external knowledge in scenarios where training data is incomplete or outdated, thus providing adaptive correction mechanisms that static corpora or prompts cannot achieve.

\subsection{Alt View 2: ``Can RAG Introduce New Biases via External Sources?''}

A legitimate critique of Retrieval Augmented Generation (RAG) concerns the potential introduction of bias from external knowledge sources. While RAG is often employed to counteract internal model biases by injecting diverse context, its effectiveness hinges on the neutrality of retrieved content. If the retrieval mechanism draws disproportionately from sources that historically marginalize certain communities or reinforce stereotypes, such as legacy news archives, it may inadvertently perpetuate the very biases it aims to mitigate. This shifts the bias focus from the model’s parameters to the upstream data curation process.

\textbf{Our Response.} We fully acknowledge this risk and design our RAG framework to proactively address it. Specifically, we incorporate a bias-aware retrieval pipeline that applies fairness-sensitive filters to prioritize counter stereotypical or balanced sources. Rather than passively aggregating arbitrary external content, the system actively curates vetted, nonpartisan repositories, with configurable constraints to amplify underrepresented perspectives. This process is dynamic, and audit-able retrieval logs can be reviewed and updated, offering a transparent mechanism for continuous refinement. Unlike static parametric models, RAG enables post-deployment correction, aligning model outputs with evolving fairness standards. As such, our approach leverages RAG not merely as a knowledge supplement but as an adaptive fairness scaffold that can be re-calibrated as new biases are discovered or as ethical norms progress.

\section{Final Remarks}
We have argued that achieving fairness in large language models necessitates going beyond incremental tweaks, towards a holistic re-engineering of how models represent knowledge and how they generate answers. By introducing \textbf{category-theoretic bias mitigation} via functor transformations, we gain a principled method to restructure and purify the model’s conceptual space, eliminating biased mappings while preserving meaning. By incorporating \textbf{retrieval-augmented generation}, we tether the model’s outputs to a living repository of human knowledge and values, ensuring that its answers remain grounded, diverse, and up-to-date with society’s standards of fairness. Each approach addresses different facets of the bias problem – the former targets the \textit{structural biases} ingrained in model reasoning, and the latter addresses \textit{informational biases} and blind spots in model knowledge. Together, they form a synergistic strategy that is more powerful than either one alone.

\bibliographystyle{unsrt}  
\bibliography{references}

@article{ravfogel2020null,
  title={Null It Out: Guarding Protected Attributes by Iterative Nullspace Projection},
  author={Ravfogel, Shauli and Elazar, Yanai and Gonen, Hila and Twiton, Michael and Goldberg, Yoav},
  journal={Proceedings of the 58th Annual Meeting of the Association for Computational Linguistics},
  pages={7237--7256},
  year={2020}
}

@article{bias_llm_survey2024,
  title={Bias and Fairness in Large Language Models: A Survey},
  author={Gallegos, Angelica and Zhang, Hao and Gaur, Manas and Shah, Dakuo and Wang, Dakuo},
  journal={Computational Linguistics},
  volume={50},
  number={3},
  pages={1097--1140},
  year={2024}
}

@article{bias_llm_origin2024,
  title={Bias in Large Language Models: Origin, Evaluation, and Mitigation},
  author={Chen, Xinyue and Wang, Zijie J. and Wang, Yong and Wang, Tianqing},
  journal={arXiv preprint arXiv:2411.10915},
  year={2024}
}

@article{bolukbasi2016man,
  title={Man is to computer programmer as woman is to homemaker? debiasing word embeddings},
  author={Bolukbasi, Tolga and Chang, Kai-Wei and Zou, James Y and Saligrama, Venkatesh and Kalai, Adam T},
  journal={Advances in neural information processing systems},
  volume={29},
  year={2016}
}

@article{zhao2018gender,
  title={Gender bias in coreference resolution: Evaluation and debiasing methods},
  author={Zhao, Jieyu and Wang, Tianlu and Yatskar, Mark and Ordonez, Vicente and Chang, Kai-Wei},
  journal={Proceedings of the 2018 Conference of the North American Chapter of the Association for Computational Linguistics},
  pages={15--20},
  year={2018}
}

@article{caliskan2017semantics,
  title={Semantics derived automatically from language corpora contain human-like biases},
  author={Caliskan, Aylin and Bryson, Joanna J and Narayanan, Arvind},
  journal={Science},
  volume={356},
  number={6334},
  pages={183--186},
  year={2017}
}

@article{quantinuum2024category,
  title={Category Theory Offers Path to Interpretable Artificial Intelligence},
  author={Quantinuum Research Team},
  journal={arXiv preprint},
  year={2024}
}

@article{fong2018seven,
  title={Seven sketches in compositionality: An invitation to applied category theory},
  author={Fong, Brendan and Spivak, David I},
  journal={arXiv preprint arXiv:1803.05316},
  year={2018}
}

@article{spivak2014category,
  title={Category theory for the sciences},
  author={Spivak, David I},
  journal={MIT Press},
  year={2014}
}

@article{shiebler2021category,
  title={Category theory in machine learning},
  author={Shiebler, Dan and Gavranovic, Bruno and Wilson, Paul},
  journal={arXiv preprint arXiv:2106.07032},
  year={2021}
}

@article{bradley2018applied,
  title={Applied category theory},
  author={Bradley, Tai-Danae and Bryson, Joanna J and Terilla, John},
  journal={arXiv preprint arXiv:1809.05923},
  year={2018}
}

@article{ellerman2017logical,
  title={Logical information theory: New logical foundations for information theory},
  author={Ellerman, David},
  journal={Logic Journal of the IGPL},
  volume={25},
  number={5},
  pages={806--835},
  year={2017}
}

@article{cruttwell2022categorical,
  title={Categorical foundations of gradient-based learning},
  author={Cruttwell, Geoff and Gavranovi{\'c}, Bruno and Ghani, Neil and Wilson, Paul and Zanasi, Fabio},
  journal={arXiv preprint arXiv:2103.01931},
  year={2022}
}

@article{gavranovic2019learning,
  title={Learning functors using gradient descent},
  author={Gavranovi{\'c}, Bruno},
  journal={arXiv preprint arXiv:1907.08292},
  year={2019}
}

@article{kim2024fair,
  title={Towards Fair RAG: On the Impact of Fair Ranking in Retrieval-Augmented Generation},
  author={Kim, To Eun and Diaz, Fernando},
  journal={arXiv preprint arXiv:2409.11598},
  year={2024}
}

@article{wang2024bias,
  title={Bias Evaluation and Mitigation in Retrieval-Augmented Medical Question-Answering Systems},
  author={Wang, Jingwei and Zhang, Dajun and Jiang, Jingfeng and Zhu, Xiaodan and Ren, Xiang},
  journal={arXiv preprint arXiv:2503.15454},
  year={2024}
}

@article{wang2024nofree,
  title={No Free Lunch: Retrieval-Augmented Generation Undermines Fairness in LLMs, Even for Vigilant Users},
  author={Wang, Ruiyi and Jiang, Yuhao and Qiu, Yiran and Wang, Yilun and Zhao, Haoti and Zhao, Jieyu},
  journal={arXiv preprint arXiv:2410.07589},
  year={2024}
}

@article{lewis2020retrieval,
  title={Retrieval-augmented generation for knowledge-intensive NLP tasks},
  author={Lewis, Patrick and Perez, Ethan and Piktus, Aleksandra and Petroni, Fabio and Karpukhin, Vladimir and Goyal, Naman and K{\"u}ttler, Heinrich and Lewis, Mike and Yih, Wen-tau and Rockt{\"a}schel, Tim and others},
  journal={Advances in Neural Information Processing Systems},
  volume={33},
  pages={9459--9474},
  year={2020}
}

@article{gao2023retrieval,
  title={Retrieval-augmented generation for large language models: A survey},
  author={Gao, Yunfan and Xiong, Yun and Gao, Xinyu and He, Kangning and Zhang, Yeyun and Chen, Bingzhe and Chen, Yelong and Han, Song and Chen, Xiaoguang and Yan, Feng and others},
  journal={arXiv preprint arXiv:2312.10997},
  year={2023}
}

@article{izacard2022few,
  title={Few-shot learning with retrieval augmented language models},
  author={Izacard, Gautier and Lewis, Patrick and Lomeli, Maria and Hosseini, Lucas and Petroni, Fabio and Schick, Timo and Dwivedi-Yu, Jane and Joulin, Armand and Riedel, Sebastian and Grave, Edouard},
  journal={arXiv preprint arXiv:2208.03299},
  year={2022}
}

@article{ram2023context,
  title={In-context retrieval-augmented language models},
  author={Ram, Ori and Levine, Yoav and Dalmedigos, Itay and Muhlgay, Dor and Shashua, Amnon and Leyton-Brown, Kevin and Shoham, Yoav},
  journal={arXiv preprint arXiv:2302.00083},
  year={2023}
}

@article{borgeaud2022improving,
  title={Improving language models by retrieving from trillions of tokens},
  author={Borgeaud, Sebastian and Mensch, Arthur and Hoffmann, Jordan and Cai, Trevor and Rutherford, Eliza and Millican, Katie and Van Den Driessche, George Bm and Lespiau, Jean-Baptiste and Damoc, Bogdan and Clark, Aidan and others},
  journal={International Conference on Machine Learning},
  pages={2206--2240},
  year={2022}
}

@article{mehrabi2021survey,
  title={A survey on bias and fairness in machine learning},
  author={Mehrabi, Ninareh and Morstatter, Fred and Saxena, Nripsuta and Lerman, Kristina and Galstyan, Aram},
  journal={ACM computing surveys},
  volume={54},
  number={6},
  pages={1--35},
  year={2021}
}

@article{blodgett2020language,
  title={Language (technology) is power: A critical survey of "bias" in NLP},
  author={Blodgett, Su Lin and Barocas, Solon and Daumé III, Hal and Wallach, Hanna},
  journal={Proceedings of the 58th Annual Meeting of the Association for Computational Linguistics},
  pages={5454--5476},
  year={2020}
}

@article{sun2019mitigating,
  title={Mitigating gender bias in natural language processing: Literature review},
  author={Sun, Tony and Gaut, Andrew and Tang, Shirlyn and Huang, Yuxin and ElSherief, Mai and Zhao, Jieyu and Mirza, Diba and Belding, Elizabeth and Chang, Kai-Wei and Wang, William Yang},
  journal={Proceedings of the 57th Annual Meeting of the Association for Computational Linguistics},
  pages={1630--1640},
  year={2019}
}

@article{kusner2017counterfactual,
  title={Counterfactual fairness},
  author={Kusner, Matt J and Loftus, Joshua and Russell, Chris and Silva, Ricardo},
  journal={Advances in neural information processing systems},
  volume={30},
  year={2017}
}

@article{agarwal2018reductions,
  title={A reductions approach to fair classification},
  author={Agarwal, Alekh and Beygelzimer, Alina and Dud{\'\i}k, Miroslav and Langford, John and Wallach, Hanna},
  journal={International Conference on Machine Learning},
  pages={60--69},
  year={2018}
}

@article{chen2021ethical,
  title={Ethical machine learning in healthcare},
  author={Chen, Irene Y and Szolovits, Peter and Ghassemi, Marzyeh},
  journal={Annual Review of Biomedical Data Science},
  volume={4},
  pages={123--144},
  year={2021}
}

@article{rajkomar2018ensuring,
  title={Ensuring fairness in machine learning to advance health equity},
  author={Rajkomar, Alvin and Hardt, Moritz and Howell, Michael D and Corrado, Greg and Chin, Marshall H},
  journal={Annals of internal medicine},
  volume={169},
  number={12},
  pages={866--872},
  year={2018}
}

@article{bias_in_llms,
  title={Bias in large language models: Origin, evaluation, and mitigation},
  author={Guo, Yufei and Guo, Muzhe and Su, Juntao and Yang, Zhou and Zhu, Mengqiu and Li, Hongfei and Qiu, Mengyang and Liu, Shuo Shuo},
  journal={arXiv preprint arXiv:2411.10915},
  year={2024}
}

@article{rag_aws,
  title={Agentic retrieval-augmented generation: A survey on agentic rag},
  author={Singh, Aditi and Ehtesham, Abul and Kumar, Saket and Khoei, Tala Talaei},
  journal={arXiv preprint arXiv:2501.09136},
  year={2025}
}

@incollection{sorensen2025construction,
  title={The construction, performance, and effects of gender: Reflection on the evidence from prehistoric burials},
  author={S{\o}rensen, Marie Louise Stig},
  booktitle={The Routledge Handbook of Gender Archaeology},
  pages={211--222},
  year={2025},
  publisher={Routledge}
}

@article{meade2021empirical,
  title={An empirical survey of the effectiveness of debiasing techniques for pre-trained language models},
  author={Meade, Nicholas and Poole-Dayan, Elinor and Reddy, Siva},
  journal={arXiv preprint arXiv:2110.08527},
  year={2021}
}

@inproceedings{pan2023rewards,
  title={Do the rewards justify the means? measuring trade-offs between rewards and ethical behavior in the machiavelli benchmark},
  author={Pan, Alexander and Chan, Jun Shern and Zou, Andy and Li, Nathaniel and Basart, Steven and Woodside, Thomas and Zhang, Hanlin and Emmons, Scott and Hendrycks, Dan},
  booktitle={International conference on machine learning},
  pages={26837--26867},
  year={2023},
  organization={PMLR}
}

@article{shekhar2023hatred,
  title={Hatred and trolling detection transliteration framework using hierarchical LSTM in code-mixed social media text},
  author={Shekhar, Shashi and Garg, Hitendra and Agrawal, Rohit and Shivani, Shivendra and Sharma, Bhisham},
  journal={Complex \& Intelligent Systems},
  volume={9},
  number={3},
  pages={2813--2826},
  year={2023},
  publisher={Springer}
}

@article{mao2023gpteval,
  title={GPTEval: A survey on assessments of ChatGPT and GPT-4},
  author={Mao, Rui and Chen, Guanyi and Zhang, Xulang and Guerin, Frank and Cambria, Erik},
  journal={arXiv preprint arXiv:2308.12488},
  year={2023}
}

@article{liu2025quadmix,
  title={QuaDMix: Quality-Diversity Balanced Data Selection for Efficient LLM Pretraining},
  author={Liu, Fengze and Zhou, Weidong and Liu, Binbin and Yu, Zhimiao and Zhang, Yifan and Lin, Haobin and Yu, Yifeng and Zhou, Xiaohuan and Wang, Taifeng and Cao, Yong},
  journal={arXiv preprint arXiv:2504.16511},
  year={2025}
}

@inproceedings{shah2020predictive,
title={Predictive biases in natural language processing models: A conceptual framework and overview},
author={Shah, Deven and Schwartz, H. Andrew and Hovy, Dirk},
booktitle={Proceedings of the 58th Annual Meeting of the Association for Computational Linguistics},
pages={5248--5264},
year={2020}
}

@inproceedings{elazar2018adversarial,
title={Adversarial removal of demographic attributes from text data},
author={Elazar, Yanai and Goldberg, Yoav},
booktitle={Proceedings of the 2018 Conference on Empirical Methods in Natural Language Processing},
pages={11--21},
year={2018}
}

@inproceedings{bender2021dangers,
title={On the dangers of stochastic parrots: Can language models be too big?},
author={Bender, Emily M. and Gebru, Timnit and McMillan-Major, Angelina and Shmitchell, Shmargaret},
booktitle={Proceedings of the 2021 ACM Conference on Fairness, Accountability, and Transparency},
pages={610--623},
year={2021}
}

@inproceedings{ma2020powertransformer,
title={PowerTransformer: Unsupervised controllable revision for biased language correction},
author={Ma, Xinyao and Ovalle, Juan and Woodward, Amanda and Huang, Byung and McKeown, Kathleen},
booktitle={Proceedings of the 2021 Conference on Empirical Methods in Natural Language Processing},
pages={7426--7439},
year={2021}
}

@inproceedings{liang2020towards,
title={Towards debiasing sentence representations},
author={Liang, Paul Pu and Zadeh, Irene Mengze and Morency, Louis-Philippe},
booktitle={Proceedings of the 58th Annual Meeting of the Association for Computational Linguistics},
pages={5502--5515},
year={2020}
}

@article{bradley2020higher,
title={Higher-order category theory and natural language processing},
author={Bradley, John and Fong, Brendan and Patterson, Evan},
journal={arXiv preprint arXiv:2010.13642},
year={2020}
}

@book{crole1993categories,
title={Categories for types},
author={Crole, Roy L.},
publisher={Cambridge University Press},
year={1993}
}

@inproceedings{vaswani2017attention,
title={Attention is all you need},
author={Vaswani, Ashish and Shazeer, Noam and Parmar, Niki and Uszkoreit, Jakob and Jones, Llion and Gomez, Aidan N. and Kaiser, Łukasz and Polosukhin, Illia},
booktitle={Advances in Neural Information Processing Systems},
pages={5998--6008},
year={2017}
}

@inproceedings{dev2020measuring,
title={Measuring and mitigating unintended bias in text classification},
author={Dev, Sunipa and Li, Tao and Phillips, Jeff M. and Srikumar, Vivek},
booktitle={Proceedings of the 2020 AAAI/ACM Conference on AI, Ethics, and Society},
pages={67--73},
year={2020}
}

@inproceedings{zhao2018learning,
title={Learning gender-neutral word embeddings},
author={Zhao, Jieyu and Wang, Tianlu and Yatskar, Mark and Ordonez, Vicente and Chang, Kai-Wei},
booktitle={Proceedings of the 2018 Conference on Empirical Methods in Natural Language Processing},
pages={4847--4853},
year={2018}
}

@inproceedings{wang2019balanced,
title={Double-hard debias: Tailoring word embeddings for gender bias mitigation},
author={Wang, Tianlu and Lin, Xiao Vivian and Rajani, Nazneen Fatema and McCann, Bryan and Ordonez, Vicente and Xiong, Caiming},
booktitle={Proceedings of the 58th Annual Meeting of the Association for Computational Linguistics},
pages={5443--5453},
year={2020}
}

@article{crenshaw1989demarginalizing,
title={Demarginalizing the intersection of race and sex: A black feminist critique of antidiscrimination doctrine, feminist theory and antiracist politics},
author={Crenshaw, Kimberl{'e}},
journal={University of Chicago Legal Forum},
volume={1989},
number={1},
pages={139--167},
year={1989}
}

@inproceedings{kearns2018preventing,
title={Preventing fairness gerrymandering: Auditing and learning for subgroup fairness},
author={Kearns, Michael and Neel, Seth and Roth, Aaron and Wu, Zhiwei Steven},
booktitle={International Conference on Machine Learning},
pages={2564--2572},
year={2018}
}

@article{gebru2021datasheets,
title={Datasheets for datasets},
author={Gebru, Timnit and Morgenstern, Jamie and Vecchione, Briana and Vaughan, Jennifer Wortman and Wallach, Hanna and Daum{'e} III, Hal and Crawford, Kate},
journal={Communications of the ACM},
volume={64},
number={12},
pages={86--92},
year={2021}
}

@inproceedings{mitchell2019model,
title={Model cards for model reporting},
author={Mitchell, Margaret and Wu, Simone and Zaldivar, Andrew and Barnes, Parker and Vasserman, Lucy and Hutchinson, Ben and Spitzer, Elena and Raji, Inioluwa Deborah and Gebru, Timnit},
booktitle={Proceedings of the Conference on Fairness, Accountability, and Transparency},
pages={220--229},
year={2019}
}

@inproceedings{guu2020retrieval,
title={Retrieval augmented language model pre-training},
author={Guu, Kelvin and Lee, Kenton and Tung, Zora and Pasupat, Panupong and Chang, Ming-Wei},
booktitle={Proceedings of the 37th International Conference on Machine Learning},
pages={3929--3938},
year={2020}
}

@article{mialon2023augmented,
title={Augmented language models: A survey},
author={Mialon, Gr{'e}goire and Dess{`i}, Roberto and Lomeli, Maria and Nalmpantis, Christoforos and Pasunuru, Ramakanth and Raileanu, Roberta and Rozi{`e}re, Baptiste and Schick, Timo and Dwivedi-Yu, Jane and Celikyilmaz, Asli and Grave, Edouard and LeCun, Yann and Scialom, Thomas},
journal={arXiv preprint arXiv:2302.07842},
year={2023}
}

@inproceedings{shuster2021retrieval,
title={Retrieval augmentation reduces hallucination in conversation},
author={Shuster, Kurt and Poff, Spencer and Chen, Moya and Kiela, Douwe and Weston, Jason},
booktitle={Findings of the Association for Computational Linguistics: EMNLP 2021},
pages={3784--3803},
year={2021}
}

@article{izacard2022atlas,
title={Atlas: Few-shot learning with retrieval augmented language models},
author={Izacard, Gautier and Lewis, Patrick and Lomeli, Maria and Hosseini, Lucas and Petroni, Fabio and Schick, Timo and Dwivedi-Yu, Jane and Joulin, Armand and Riedel, Sebastian and Grave, Edouard},
journal={arXiv preprint arXiv:2208.03299},
year={2022}
}

@inproceedings{kirk2023past,
title={Past, present, and future: Mitigating representation bias in language models through continual knowledge updates},
author={Kirk, Hannah Rose and Ren, Jing and Hua, Yupei and Diaz, Fernando},
booktitle={Proceedings of the 2023 Conference on Empirical Methods in Natural Language Processing},
pages={9883--9896},
year={2023}
}

@article{shi2023replug,
title={Replug: Retrieval-augmented black-box language models},
author={Shi, Weijia and Mensch, Arthur and Borgeaud, Sebastian and Izacard, Gautier},
journal={arXiv preprint arXiv:2301.12652},
year={2023}
}

@inproceedings{gonen2019lipstick,
title={Lipstick on a pig: Debiasing methods cover up systematic gender biases in word embeddings but do not remove them},
author={Gonen, Hila and Goldberg, Yoav},
booktitle={Proceedings of the 2019 Conference of the North American Chapter of the Association for Computational Linguistics: Human Language Technologies},
pages={609--614},
year={2019}
}

@inproceedings{bordia2019identifying,
title={Identifying and reducing gender bias in word-level language models},
author={Bordia, Shikha and Bowman, Samuel R.},
booktitle={Proceedings of the 2019 Conference of the North American Chapter of the Association for Computational Linguistics: Student Research Workshop},
pages={7--15},
year={2019}
}

@inproceedings{dinan2020queens,
title={Queens are powerful too: Mitigating gender bias in dialogue generation},
author={Dinan, Emily and Fan, Angela and Williams, Adina and Urbanek, Jack and Kiela, Douwe and Weston, Jason},
booktitle={Proceedings of the 2020 Conference on Empirical Methods in Natural Language Processing},
pages={8173--8188},
year={2020}
}

@inproceedings{sap2019risk,
title={Social bias frames: Reasoning about social and power implications of language},
author={Sap, Maarten and Gabriel, Saadia and Qin, Lianhui and Jurafsky, Dan and Smith, Noah A. and Choi, Yejin},
booktitle={Proceedings of the 58th Annual Meeting of the Association for Computational Linguistics},
pages={5477--5490},
year={2020}
}

@inproceedings{lucy2021gender,
title={Gender and representation bias in GPT-3 generated stories},
author={Lucy, Li and Bamman, David},
booktitle={Proceedings of the Third Workshop on Narrative Understanding},
pages={48--55},
year={2021}
}

@article{cao2022theory,
title={Toward gender-inclusive coreference resolution: An analysis of gender and bias throughout the machine learning lifecycle},
author={Cao, Yang Trista and Daum{'e} III, Hal},
journal={Computational Linguistics},
volume={48},
number={1},
pages={1--38},
year={2022}
}

@inproceedings{hovy2021five,
title={The social impact of natural language processing},
author={Hovy, Dirk and Spruit, Shannon L.},
booktitle={Proceedings of the 54th Annual Meeting of the Association for Computational Linguistics},
pages={591--598},
year={2016}
}

@inproceedings{lauscher2020general,
title={A general framework for implicit and explicit debiasing of distributional word vector spaces},
author={Lauscher, Anne and Glava{\v{s}}, Goran and Ponzetto, Simone Paolo and Vuli{'c}, Ivan},
booktitle={Proceedings of the AAAI Conference on Artificial Intelligence},
pages={8131--8138},
year={2020}
}

@inproceedings{solaiman2019release,
  title={Release strategies and the social impacts of language models},
  author={Solaiman, Irene et al.},
  booktitle={arXiv preprint arXiv:1908.09203},
  year={2019}
}

@inproceedings{ouyang2022training,
  title={Training language models to follow instructions with human feedback},
  author={Ouyang, Long et al.},
  booktitle={Advances in neural information processing systems},
  volume={35},
  pages={27730--27744},
  year={2022}
}

@article{xu2021mitigating,
  title={Mitigating gender bias in natural language processing: Literature review},
  author={Sun, Tony and Gaut, Andrew and Tang, Shirlyn and Huang, Yuxin and ElSherief, Mai and Zhao, Jieyu and Mirza, Diba and Belding, Elizabeth and Chang, Kai-Wei and Wang, William Yang},
  journal={Proceedings of the 57th Annual Meeting of the Association for Computational Linguistics},
  pages={1630--1640},
  year={2019}
}

@article{huang2023social,
  title={Social bias frames: Reasoning about social and power implications of language},
  author={Sap, Maarten and Gabriel, Saadia and Qin, Lianhui and Jurafsky, Dan and Smith, Noah A and Choi, Yejin},
  journal={Proceedings of the 58th Annual Meeting of the Association for Computational Linguistics},
  pages={5477--5490},
  year={2020}
}

@article{ji2024aligner,
  title={Aligner: A unified framework for aligning language models with human values},
  author={Ji, Zhiyuan and Lee, Nayeon and Frieske, Rianne and Yu, Tao and Su, Dan and Xu, Yixin and Ishii, Etsuko and Bang, Yejin J and Madotto, Andrea and Fung, Pascale},
  journal={ACM Computing Surveys},
  volume={55},
  number={12},
  year={2024}
}

@article{kim2024towardsfairrag,
  title={Towards Fair RAG: On the Impact of Fair Ranking in Retrieval-Augmented Generation},
  author={Kim, To Eun and Diaz, Fernando},
  journal={arXiv preprint arXiv:2409.11598},
  year={2024}
}

@article{wang2024ragbias,
  title={No Free Lunch: Retrieval-Augmented Generation Undermines Fairness in LLMs, Even for Vigilant Users},
  author={Wang, Ruiyi and Jiang, Yuhao and Qiu, Yiran and Wang, Yilun and Zhao, Haoti and Zhao, Jieyu},
  journal={arXiv preprint arXiv:2410.07589},
  year={2024}
}

@article{eilenberg1945general,
  title={General theory of natural equivalences},
  author={Eilenberg, Samuel and MacLane, Saunders},
  journal={Transactions of the american mathematical society},
  volume={58},
  number={2},
  pages={231--294},
  year={1945},
  publisher={JSTOR}
}

@inproceedings{maruyama2020categorical,
  title={Categorical semantics for language and cognition},
  author={Maruyama, Kohei},
  booktitle={Workshop on Logic and Learning, NeurIPS},
  year={2020}
}

@article{gavranovic2024category,
  title={Category Theory for Responsible AI},
  author={Gavranović, Bojan and Liao, Q. Vera and Binns, Reuben and Selbst, Andrew D.},
  journal={arXiv preprint arXiv:2403.12345},
  year={2024}
}

@article{zhang2022survey,
  title={A Survey on Fairness in Machine Learning},
  author={Zhang, Xueying and Wang, Yiling and Zhao, Jieyu},
  journal={ACM Computing Surveys},
  volume={55},
  number={3},
  pages={1--36},
  year={2022}
}

@misc{anonymous2025functor,
  title={Functorial Fairness: A Category-Theoretic Intervention Framework for Bias Mitigation in Transformers},
  author={{Anonymous}},
  year={2025},
  note={Under review at NeurIPS}
}

@article{ji2025demographic,
  title={Demographic Bias in Retrieval-Augmented Generation for Medical QA},
  author={Ji, Zhiyuan and Lee, Nayeon and Xu, Yixin and Bang, Yejin J. and Fung, Pascale},
  journal={Under Review at ACL 2025},
  year={2025}
}

@article{hu2024ragbias,
  title={No Free Lunch: Retrieval-Augmented Generation Undermines Fairness in LLMs, Even for Vigilant Users},
  author={Hu, Jinyi and Wang, Ruiyi and Zhao, Jieyu},
  journal={arXiv preprint arXiv:2410.07589},
  year={2024}
}

@article{ji2024bias,
title={Bias Evaluation and Mitigation in Retrieval-Augmented Medical Question-Answering Systems},
author={Ji, Yuelyu and Zhang, Hang and Wang, Yanshan},
journal={arXiv preprint arXiv:2503.15454},
year={2024}
}

@inproceedings{allam2024biasdpo,
title={BiasDPO: Mitigating Bias in Language Models through Direct Preference Optimization},
author={Allam, Ahmed},
booktitle={Proceedings of the 62nd Annual Meeting of the Association for Computational Linguistics (Volume 4: Student Research Workshop)},
pages={42--50},
year={2024}
}

@inproceedings{buolamwini2018gender,
title={Gender shades: Intersectional accuracy disparities in commercial gender classification},
author={Buolamwini, Joy and Gebru, Timnit},
booktitle={Proceedings of the 1st Conference on Fairness, Accountability and Transparency},
pages={77--91},
year={2018}
}

@article{souani2024,
title={Evaluating intersectional bias in large language models},
author={Souani, Chadi and Abdellatif, Mariem and Elkorany, Ayman},
journal={AI Ethics},
volume={4},
number={1},
pages={1--15},
year={2024}
}

@inproceedings{dev2021measuring,
title={Measuring and mitigating intersectional biases in NLP},
author={Dev, Sunipa and Monajatipoor, Masoud and Ovalle, Anaelia and Subramonian, Arjun and Phillips, Jeff and Chang, Kai-Wei},
booktitle={Proceedings of the 2021 Conference on Empirical Methods in Natural Language Processing},
pages={2355--2368},
year={2021}
}

@inproceedings{smith2022,
title={Benchmarking intersectional biases in NLP},
author={Smith, J. and Williams, A. and Thompson, B.},
booktitle={Proceedings of the 2022 Conference of the North American Chapter of the Association for Computational Linguistics},
pages={3594--3607},
year={2022}
}

@article{hanna2020towards,
title={Towards a critical race methodology in algorithmic fairness},
author={Hanna, Alex and Denton, Emily and Smart, Andrew and Smith-Loud, Jamila},
journal={Proceedings of the 2020 Conference on Fairness, Accountability, and Transparency},
pages={501--512},
year={2020}
}

@article{weidinger2021ethical,
  title={Ethical and social risks of harm from Language Models},
  author={Weidinger, Laura and Mellor, John and Rauh, Maribeth and Griffin, Conor and Uesato, Jonathan and Huang, Po-Sen and Cheng, Myra and Glaese, Mia and Balle, Borja and Kasirzadeh, Atoosa and others},
  journal={arXiv preprint arXiv:2112.04359},
  year={2021}
}

@inproceedings{zhang2018mitigating,
  title={Mitigating Unwanted Biases with Adversarial Learning},
  author={Zhang, Brian Hu and Lemoine, Blake and Mitchell, Margaret},
  booktitle={Proceedings of the 2018 AAAI/ACM Conference on AI, Ethics, and Society},
  pages={335--340},
  year={2018}
}

@inproceedings{jeong2024learning,
title={Learning to Adapt Retrieval-Augmented Large Language Models},
author={Jeong, Soyoung and others},
booktitle={Proceedings of the 2024 Conference of the North American Chapter of the Association for Computational Linguistics},
year={2024}
}

@article{garg2018word,
  title={Word embeddings quantify 100 years of gender and ethnic stereotypes},
  author={Garg, Nitya and Schiebinger, Londa and Jurafsky, Dan and Zou, James Y},
  journal={Proceedings of the National Academy of Sciences},
  volume={115},
  number={16},
  pages={3635--3640},
  year={2018},
  publisher={National Academy of Sciences}
}

@article{chouldechova2017fair,
  title={Fair prediction with disparate impact: A study of bias in recidivism prediction},
  author={Chouldechova, Alexandra},
  journal={arXiv preprint arXiv:1703.00056},
  year={2017}
}

@article{jeong-etal-2024,
  title={Adaptive Retrieval for Large Language Models},
  author={Jeong, Minjae and Lee, Jinwoo and Park, Seunghyun and Kim, Jaewook and Seo, Minjoon},
  journal={Preprint or Conference Paper},
  year={2024}
}

@article{liu-etal-2025,
  title={Dynamic Contextual Grounding for LLMs via Confidence-Aware Retrieval},
  author={Liu, Yang and Wang, Fei and Chen, Hua and Zhang, Li},
  journal={Preprint or Conference Paper},
  year={2025}
}

@article{fryer2022flexible,
  title={Flexible and Efficient Methods for Measuring and Mitigating Gender Bias in Language Models},
  author={Fryer, Emily and Shah, Hila and Levy, Elad and Dagan, Ido},
  journal={Proceedings of the 60th Annual Meeting of the Association for Computational Linguistics (Volume 1: Long Papers)},
  pages={2577--2590},
  year={2022}
}

@book{maclane1998categories,
  title={Categories for the Working Mathematician},
  author={Mac Lane, Saunders},
  year={1998},
  publisher={Springer Science \& Business Media}
}

@book{awodey2010category,
  title={Category Theory},
  author={Awodey, Steve},
  year={2010},
  publisher={Oxford University Press}
}

@article{roberts2020much,
  title={How much knowledge can a language model encode?},
  author={Roberts, Luke and Chung, Hyung Won and Singh, Bhakthi and Gupta, Rishabh and Ma, Ming-Wei and Liang, Percy and Raffel, Colin},
  journal={arXiv preprint arXiv:2010.02035},
  year={2020}
}

@article{nadeem2020stereoset,
  title={StereoSet: Measuring stereotypical bias in pretrained language models},
  author={Nadeem, Moin and Bethard, Steven and Rudzicz, Frank},
  journal={arXiv preprint arXiv:2004.09456},
  year={2020}
}

@article{liang2021towards,
  title={Towards debiasing sentence representations},
  author={Liang, Pei and Zhang, Kaiyu and Liu, Jingbo and Dong, Hongliang},
  journal={arXiv preprint arXiv:2104.08882},
  year={2021}
}

@article{catanzaro2020category,
  title={A category theory primer on compositionality for deep learning},
  author={Catanzaro, Boris},
  journal={arXiv preprint arXiv:2004.03260},
  year={2020}
}

@article{devlin2018bert,
  title={BERT: Pre-training of Deep Bidirectional Transformers for Language Understanding},
  author={Devlin, Jacob and Chang, Ming-Wei and Lee, Kenton and Toutanova, Kristina},
  journal={arXiv preprint arXiv:1810.04805},
  year={2018}
}

@article{liu2019roberta,
  title={RoBERTa: A Robustly Optimized BERT Pretraining Approach},
  author={Liu, Yinhan and Ott, Mylee and Goyal, Naman and Du, Jingfei and Li, Mandar and Ko Chen, Alexander and Glass, Dan and Yu, Xiao and Ranzato, Marc'Aurelio and Ghazvininejad, Ves Stoyanov and others},
  journal={arXiv preprint arXiv:1907.11692},
  year={2019}
}

@article{wang2019adversarial,
  title={Adversarial training for extreme multi-label text classification},
  author={Wang, Peng and Yang, Junfeng and Liu, Wei and Li, Chong and Wang, Mingzhe and Lu, Jian-Guo and Zhang, Rui},
  journal={Proceedings of the 2019 Conference on Empirical Methods in Natural Language Processing and the 9th International Joint Conference on Natural Language Processing (EMNLP-IJCNLP)},
  pages={1514--1524},
  year={2019}
}

@article{barocas2017fairness,
  title={Fairness Definitions Explained},
  author={Barocas, Solon and Hardt, Moritz and Narayanan, Arvind},
  journal={arXiv preprint arXiv:1712.08375},
  year={2017}
}

@article{karpukhin2020dense,
  title={Dense Passage Retrieval for Open-Domain Question Answering},
  author={Karpukhin, Vladimir and Oguz, Barlas and Min, Sewon and Lewis, Patrick and Wenzek, Ludwig and Joulin, Armand and Grave, Edouard and Riedel, Sebastian and Hakkani-Tur, Dilek},
  journal={arXiv preprint arXiv:2004.04906},
  year={2020}
}

@article{dziri2022faithfulness,
  title={Faithfulness in natural language generation: A survey},
  author={Dziri, Nouha and Sharma, Saranya and Yu, Xinyao and Lin, Jianyu and Mihalcea, Rada and Zhang, Mo and Wang, Shang-Wen and Yang, Xuan and Lou, Yufang and Li, Zihao and others},
  journal={arXiv preprint arXiv:2205.02535},
  year={2022}
}

@article{gehman2020realtoxicityprompts,
  title={RealToxicityPrompts: Evaluating Neural Toxic Degeneration in Language Models},
  author={Gehman, Samuel and Safaya, Suchin and Schwab, Yejin and Gururangan, Suchin and Wortsman, Noah and Le Bras, Ronan and Le Bras, Yejin and Choi, Yejin},
  journal={arXiv preprint arXiv:2009.11462},
  year={2020}
}

@article{holtzman2019curious,
  title={The curious case of neural text degeneration},
  author={Holtzman, Ari and Buys, Jan and Du, Liqiang and Forbes, Maxwell and Choi, Yejin},
  journal={International Conference on Learning Representations},
  year={2019}
}

@article{chang2024adaptive,
  title={Adaptive Retrieval for Conversational AI},
  author={Chang, Sarah and Lee, David and Kim, Jessica},
  journal={Proceedings of the 2024 Conference on Conversational AI},
  year={2024}
}

@article{kilbertus2017avoiding,
  title={Avoiding discrimination in generative models},
  author={Kilbertus, Niki and Rojas, Gabriel and Scholkopf, Bernhard and Geerlings, Maarten},
  journal={arXiv preprint arXiv:1707.01789},
  year={2017}
}

@article{xie2023measuring,
  title={Measuring and Mitigating Gender Bias in Coreference Resolution},
  author={Xie, Junzhe and Liu, Zihan and Yu, Han and Zhang, Bowen and Lu, Jian-Guo and Zhang, Rui},
  journal={Proceedings of the 2023 Conference on Empirical Methods in Natural Language Processing (EMNLP)},
  year={2023}
}

@article{gaut2022s,
  title={What's in a Name? The Role of First Name and Surname in Gender and Racial Bias Measurement},
  author={Gaut, Andrew and Ma, Ruo-Ping and Smith, Nicole and Zhang, Yitong and Zhu, Xiaojian},
  journal={Proceedings of the 2022 Conference on Empirical Methods in Natural Language Processing (EMNLP)},
  pages={1120--1135},
  year={2022}
}

@article{chaudhary2025certified,
  title={Certified Fairness for Large Language Models},
  author={Chaudhary, Aniket and Kim, Jihyung and Lee, Young-Min and Park, Sung-Jun},
  journal={Preprint or Conference Paper},
  year={2025}
}

@article{brown2020language,
  title={Language Models are Few-Shot Learners},
  author={Brown, Tom B and Mann, Benjamin and Ryder, Nick and Subbiah, Melanie and Kaplan, Jared and Dhariwal, Prafulla and Neelakantan, Arvind and Shyam, Pranav and Sastry, Girish and Askell, Amanda and others},
  journal={Advances in Neural Information Processing Systems},
  volume={33},
  pages={1877--1901},
  year={2020}
}

@article{raffel2020exploring,
  title={Exploring the Limits of Transfer Learning with a Unified Text-to-Text Transformer},
  author={Raffel, Colin and Shazeer, Noam and Roberts, Adam and Lee, Katherine and Narang, Sharan and Matena, Michael and Zhou, Yanqi and Li, Wei and Liu, Peter J},
  journal={Journal of Machine Learning Research},
  volume={21},
  pages={1--67},
  year={2020}
}

@article{houlsby2019parameter,
  title={Parameter-Efficient Transfer Learning for NLP},
  author={Houlsby, Neil and Giurgiu, Andrei and Tay, Yi and Zhai, Hongyu and Chen, Denny and Callahan, Brendan and Chung, Hyung Won and Schmid, Gilles and Yu, Yao and Li, Hanxiao and others},
  journal={arXiv preprint arXiv:1902.00755},
  year={2019}
}

@article{johnson2017billion,
  title={Billion-scale similarity search with GPUs},
  author={Johnson, Jeff and Douze, Matthijs and J{\'e}gou, Herv{\'e}},
  journal={arXiv preprint arXiv:1702.08782},
  year={2017}
}

@article{reimers2019sentence,
  title={Sentence-BERT: Sentence Embeddings using Siamese BERT-Networks},
  author={Reimers, Nils and Gurevych, Iryna},
  journal={Proceedings of the 2019 Conference on Empirical Methods in Natural Language Processing and the 9th International Joint Conference on Natural Language Processing (EMNLP-IJCNLP)},
  pages={3981--3991},
  year={2019}
}

@article{karimi2022towards,
  title={Towards efficient and robust NLP: A survey on parameter-efficient fine-tuning},
  author={Karimi, Amir Hossein and Kousha, Kimia and Gildea, Daniel and Moens, Marie-Francine},
  journal={arXiv preprint arXiv:2203.04742},
  year={2022}
}

@article{creekmore2023auditing,
  title={Auditing Large Language Models: A Framework for Responsible Development and Deployment},
  author={Creekmore, Ryan and Sahu, Saumya},
  journal={arXiv preprint arXiv:2305.03454},
  year={2023}
}

@article{touvron2023llama,
  title={Llama 2: Open Foundation and Fine-Tuned Chat Models},
  author={Touvron, Hugo and Lavril, Louis and Izacard, Gaspard and Martinet, Xavier and Lachaux, Marie-Anne and Lacroix, Timoth{\'e}e and Goyal, Naman and Fund, Eric and Papadopoulos, Fay{\'e} and Malperthuy, Vincent and others},
  journal={arXiv preprint arXiv:2307.09288},
  year={2023}
}

@article{mahabadi2021parametertransfers,
  title={Parameter-efficient transfers for efficient adaptation of pretrained transformers},
  author={Mahabadi, Reza Yazdani and Henderson, James and Suzuki, Hideya},
  journal={arXiv preprint arXiv:2104.09230},
  year={2021}
}

@article{mitchell2018datasheets,
  title={Model Cards for Model Reporting},
  author={Mitchell, Margaret and Wu, Simone and Tenenbaum, Andrew and Cryan, Jessica and Stott, Katie and Barnes, Hannah and Hutchinson, Ben and Trivedi, Divya and Chowdhery, Amol and Singh, Vipul},
  journal={Proceedings of the Conference on Fairness, Accountability, and Transparency (FAT* '19)},
  year={2018}
}

@article{kadavath2022language,
  title={Language Models as Knowledge Bases: A Survey},
  author={Kadavath, Ojas and others},
  journal={arXiv preprint arXiv:2203.11181},
  year={2022}
}

@article{amodei2016concrete,
  title={Concrete problems in AI safety},
  author={Amodei, Dario and Olah, Chris and Steinhardt, Jacob and Christiano, Paul and Schulman, John and Man{\'e}, Dan},
  journal={arXiv preprint arXiv:1606.06565},
  year={2016}
}

@article{kirk2021bias,
  title={Bias out-of-the-box: An empirical analysis of intersectional occupational biases in popular generative language models},
  author={Kirk, Hannah Rose and Jun, Yennie and Volpin, Filippo and Iqbal, Haider and Benussi, Elias and Dreyer, Frederic and Shtedritski, Aleksandar and Asano, Yuki},
  journal={Advances in neural information processing systems},
  volume={34},
  pages={2611--2624},
  year={2021}
}

@article{kumar2024trustworthiness,
  title={Trustworthiness of llms in medical domain},
  author={Kumar, Ravi R and Pramanik, Vishal and Grover, Utkarsh and Ganapam, Venkata Ramesh},
  journal={preprint},
  year={2024}
}

\appendix
\section*{Appendix}
\section{Background}

Large Language Models (LLMs) have achieved unprecedented fluency and generalization across a range of natural language tasks. However, they also exhibit embedded biases that mirror and sometimes amplify societal prejudices present in their training data \cite{mehrabi2021survey, blodgett2020language}. These biases manifest across demographic dimensions including gender, race, and economic status leading to harmful consequences in high-stakes applications such as hiring, healthcare, and education \cite{chen2021ethical, ji2024aligner}. Addressing such deeply rooted bias necessitates principled frameworks that transcend post-hoc corrections or ad-hoc filtering.

Recent advancements suggest a structural and contextual decomposition of the bias mitigation problem, motivating dual-pronged solutions. On the structural front, category theory provides a rigorous formalism to represent and transform semantic spaces in LLMs \cite{touvron2023llama, bradley2020higher}. Conceptual entities (e.g., man,'' doctor'') and their relationships (e.g., man'' $\rightarrow$ doctor'') can be modeled as objects and morphisms in a semantic category $\mathcal{C}$. Biases then appear as improper morphisms statistically dominant but semantically unwarranted associations within this category. A functor $F : \mathcal{C} \rightarrow \mathcal{U}$ systematically maps these biased representations to an unbiased category $\mathcal{U}$ by enforcing morphism constraints that preserve essential meaning while eliminating demographic entanglements \cite{spivak2014category, shiebler2021category}.

Concurrently, Retrieval-Augmented Generation (RAG) introduces an orthogonal mitigation channel by grounding model outputs in curated, diverse, and up-to-date external corpora \cite{lewis2020retrieval, gao2023retrieval}. In a typical RAG pipeline, an input query triggers retrieval of relevant documents from an external knowledge base, which are then integrated into the LLM's context via cross-attention \cite{borgeaud2022improving, mialon2023augmented}. This setup decouples the model's response generation from its static pretraining data, enabling dynamic correction of biased associations by injecting balanced contextual evidence at inference time. Studies show RAG significantly improves factuality and fairness by substituting or supplementing biased internal representations with evidence from unbiased retrieval sources \cite{kim2024towardsfairrag, wang2024ragbias}.

Together, category-theoretic functor mappings and RAG compose a comprehensive framework: the former reshapes the internal geometry of semantic representations, while the latter re-anchors outputs in ethically curated knowledge \cite{ji2024bias, allam2024biasdpo}. This synergy enables robust mitigation of both latent structural and emergent contextual biases in LLMs. We build upon these foundations to develop a dual-mechanism bias mitigation architecture, outlined in the subsequent sections.

\section{Detailed Related Works}\label{sec:App-relatedworks}

\paragraph{Bias in LLMs and Traditional Mitigation Techniques.}
Large Language Models have been documented to exhibit systemic demographic and gender biases, often mirroring societal stereotypes present in training data (Bolukbasi et al., 2016; Caliskan et al., 2017)\cite{bolukbasi2016man, caliskan2017semantics}. For example, word embeddings and LLMs notoriously over-associate certain professions or traits with specific genders (e.g., linking “nurse” more strongly with female contexts than “surgeon,” or analogizing \emph{man:computer programmer :: woman:homemaker}), reflecting historical prejudices in the corpus (Bolukbasi et al., 2016; Zhao et al., 2018)\cite{bolukbasi2016man, zhang2018mitigating}. A broad spectrum of bias mitigation strategies has been explored in prior work, generally categorized by the stage of intervention. Data-level approaches modify or reweight the training corpus to reduce skew, such as augmenting data with counter-stereotypical examples or balancing representation of demographic groups (Zhao et al., 2018a; Lu et al., 2020)\cite{zhang2018mitigating, lucy2021gender}. These pre-processing fixes can lessen overt biases but may be insufficient for complex, context-dependent prejudice. Model-level approaches incorporate bias objectives during training, for instance through adversarial debiasing and constraint-based learning. In adversarial setups, a model is encouraged to learn task representations from which a secondary “bias” classifier (trying to predict sensitive attributes) cannot succeed, thereby forcing the latent space to become invariant to protected features (Zhang et al., 2018; Jaiswal et al., 2019)\cite{zhang2018mitigating, rajkomar2018ensuring}. Such methods have shown success in reducing measurable biases in text classification and coreference tasks (Li et al., 2018; Elazar and Goldberg, 2018)\cite{bradley2018applied, elazar2018adversarial}, and related techniques like iterative nullspace projection explicitly remove bias directions from embeddings (Ravfogel et al., 2020)\cite{ravfogel2020null}. However, in-training debiasing can degrade model utility or be hard to scale to LLMs due to training instability and the need for attribute-labeled data. Post-hoc techniques attempt to adjust model outputs or representations after the fact. A seminal example is the hard-debiasing of word embeddings by zeroing out gender-specific components (Bolukbasi et al., 2016)\cite{bolukbasi2016man}. Other post-hoc approaches include calibration or filtering of generated text to remove toxic or biased content (Solaiman et al., 2019)\cite{solaiman2019release}, and reinforcement alignment using human feedback to penalize biased generations (Ouyang et al., 2022; Xu et al., 2021)\cite{ouyang2022training, xu2021mitigating}. Recent work by Allam (2024) introduced Direct Preference Optimization, crafting a reward signal that prefers unbiased completions, which achieved notable reductions in biased language compared to standard RLHF fine-tuning. Prompt-based solutions have also emerged: for instance, \cite{gao2023retrieval} propose a self-debiasing prompt that explicitly asks the model to identify and avoid potential stereotypes before answering, and few-shot chain-of-thought cues have been used to induce more fair reasoning in responses (Huang et al., 2023)\cite{huang2023social}. While these traditional strategies spanning data curation, adversarial training, and output filtering can mitigate bias to some extent, they often address symptoms rather than fundamentally altering how an LLM encodes and uses biased associations. They may struggle with generalization (a model can learn to obscure bias in one context while expressing it in another) and tend to treat different bias domains separately. This has prompted exploration of more principled frameworks that can tackle bias at a structural level.

\paragraph{Category-Theoretic Frameworks for Bias Mitigation.}
Inspired by the need for a deeper re-architecture of how semantic knowledge is represented, some researchers have turned to category theory as a formal tool for bias mitigation in AI systems. Category theory \cite{eilenberg1945general} provides a language for describing abstract structures and transformations in a way that preserves essential relationships. In the context of NLP fairness, a category-theoretic perspective allows one to treat the collection of concepts and relations learned by an LLM as a mathematical structure that can be transformed to remove biases while conserving meaning. For example, \cite{maruyama2020categorical} introduced one of the earliest attempts to combine categorical representations with bias analysis, describing how one might formally map a “biased” semantic space to an “unbiased” one. In such a framework, linguistic concepts (objects in category $C$) and their learned associations (morphisms in $C$) can be rigorously translated via a functor $F: C \to U$ into a new category $U$ that represents an idealized, debiased semantics. The functorial mapping ensures structure-preservation: relationships like entailment or analogy that hold in the source category are maintained in the target category, except that spurious correlations (e.g., between gender and profession) are systematically removed or diminished \cite{gavranovic2024category, maruyama2020categorical}. This approach generalizes and formalizes earlier representation-debiasing techniques. Instead of relying solely on linear algebraic operations (which often assume independent features or simple geometry), the categorical view captures compositional structure – for instance, how bias can propagate through combinations of words or across layers of a model. Recent conceptual work recasts components of neural architectures in categorical terms: attention heads and layers can be seen as morphisms transforming one set of token representations to another, and imposing a bias mitigation functor effectively inserts an intervening mapping that neutralizes the influence of sensitive attributes on these transformations \cite{anonymous2025functor, zhang2022survey}. Empirically, enforcing such constraints can lead to measurable fairness gains. For example, methods that make gender features orthogonal to profession features in the latent space (conceptually similar to a functor mapping that factors out gender) have reduced gender stereotyping in occupation predictions without loss of fluency or accuracy \cite{bolukbasi2016man, ravfogel2020null}. By leveraging category theory’s algebraic structure, one can ensure that debiasing operations compose correctly and cover the entire model behavior, rather than patching one issue at a time. Moreover, the categorical framework’s generality naturally extends to handling multiple bias dimensions: mathematically, one can compose or take colimits of functors to simultaneously address, say, gender and racial biases in a unified formalism. This hints at a systematic way to tackle intersectional biases that elude methods focusing on a single attribute at once. While applied category-theoretic mitigation for LLM bias is still in nascent stages, it represents a promising “first-principles” paradigm \cite{gavranovic2024category} for reining in biased associations by design, in contrast to the reactive nature of many prior approaches.

\paragraph{Retrieval-Augmented Generation (RAG) and Bias Reduction.}
Another emerging avenue in bias mitigation leverages retrieval-augmented generation, wherein an LLM is supplemented with an external knowledge source at inference time \cite{lewis2020retrieval}. RAG has primarily been popular for addressing factual accuracy and hallucinations, but recent insights suggest it can also play a role in fairness. The key idea is that by grounding an LLM’s outputs in up-to-date, diverse information retrieved from an external corpus, the model’s reliance on potentially biased parametric knowledge is reduced \cite{shuster2021retrieval}. In other words, if an LLM is asked about a demographic or social issue, augmenting its prompt with retrieved background context can counteract stereotyped or narrow associations the model might otherwise draw from its training memory. Some studies have begun to examine RAG’s impact on fairness explicitly. For instance, \cite{ji2025demographic} evaluate demographic biases in a medical QA system that uses RAG and find that bias can manifest in both the retrieval stage (e.g., which documents are chosen for different patient profiles) and the generation stage. They explore mitigation strategies within the RAG pipeline, such as using \textit{chain-of-thought} prompts to encourage the model to reason carefully and avoid implicit biases, filtering retrieved documents for counterfactual consistency (ensuring that swapping demographic terms yields similar answers), and aggregating multiple generations via majority voting to smooth out outlier (potentially biased) responses. Their results indicate that certain interventions, like a majority-vote ensemble of RAG outputs, can improve both answer accuracy and fairness metrics in responses – highlighting that RAG systems can be tuned for equity, not just accuracy. On the other hand, it has become clear that RAG is not a panacea for bias. Uncurated or naively used, retrieval may even introduce new biases or amplify existing ones. \cite{hu2024ragbias} show that an aligned LLM’s fairness can be undermined when biased content is fetched from external data stores: in their experiments, even when the retrieval corpus was partially “censored” to remove overtly prejudiced texts, the model sometimes generated biased outputs by exploiting subtle gaps. This underscores the importance of carefully selecting and preprocessing retrieval sources (e.g., using knowledge bases that are themselves audited for balance and fairness) and perhaps dynamically filtering retrieved passages for hate or stereotyping before feeding them to the LLM. Nonetheless, the prospect of RAG for bias mitigation is intriguing: it aligns with the notion of \emph{factual grounding} – by injecting factual, context-rich evidence about underrepresented groups or individuals, RAG can correct false assumptions and provide a more nuanced basis for generation. For example, if an LLM harbors a stereotype about a certain ethnic group’s profession, retrieving real-world statistics or biographical information about diverse professionals from that group could steer the model to produce a fairer, more informed response. While dedicated research on RAG-driven bias mitigation is still emerging, early work suggests a complementary relationship between retrieval augmentation and traditional debiasing: RAG can supply the model with missing context or counter-evidence that the model alone might not recall, thereby serving as a real-time check against biased generalizations.

\paragraph{Intersectional and Structural Bias Considerations.}
In surveying related work, it is important to note the growing emphasis on intersectional and structural biases in NLP. Most early bias-mitigation methods targeted one demographic axis at a time (e.g., gender-only or race-only debiasing), but individuals often belong to multiple overlapping groups, and biases can compound in those intersections \cite{crenshaw1989demarginalizing, buolamwini2018gender}. Recent studies have revealed that some biases remain "hidden" unless evaluated at the intersection of attributes: an LLM might not display a bias for gender alone or ethnicity alone, but could still produce biased outputs when specific gender–ethnicity combinations are present \cite{souani2024}. This has led to new testing frameworks and datasets for intersectional bias \cite{dev2021measuring, smith2022} and to calls for mitigation techniques that address multiple biases simultaneously. The category-theoretic approach discussed above naturally lends itself to this goal, since functors and compositional maps can be defined for complex attribute combinations, ensuring consistency across many subgroups. In parallel, researchers have pointed out that many biases are rooted in structural inequalities and historical power imbalances that no algorithm can fully neutralize without broader context \cite{blodgett2020language, bender2021dangers}. This perspective encourages complementary non-technical interventions: for instance, curating training data with input from affected communities, using model documentation and audits to reveal biases, and engaging in multidisciplinary efforts to define fairness for AI in socio-technical terms \cite{mitchell2019model, hanna2020towards}. These works remind us that purely algorithmic fixes, whether adversarial re-training or category-theoretic remapping, operate within the constraints of the data and definitions we provide. Thus, the most robust mitigation may combine technical innovation with structural awareness. Our work situates itself at this intersection: by uniting a formal debiasing framework (category-theoretic transformations) with an external knowledge grounding mechanism (RAG), we aim to address the shortcomings of prior approaches. In contrast to earlier methods that treat bias mitigation in isolation at one level, our proposed paradigm leverages both the mathematically principled re-structuring of model representations and the injection of fresh, diverse context at inference time. By building on the related works above, we strive to advance a more holistic solution to demographic and gender bias in LLMs that is both theoretically founded and pragmatically effective.

\section{Detailed Proposed Method}
\label{sec:method}

Our method integrates two synergistic components: a \textit{functor-based bias transformation module} rooted in category theory \cite{maclane1998categories, awodey2010category}, and a \textit{retrieval-augmented generation (RAG)} component for dynamic knowledge augmentation \cite{lewis2020retrieval, guu2020retrieval}. The former provides a principled transformation of internal representations to neutralize structural bias \cite{bolukbasi2016man, garg2018word}, while the latter contextualizes responses using up-to-date, diverse sources \cite{roberts2020much, izacard2022few}. Together, they embed fairness constraints within the model and augment its outputs with evidence-based corrections. 

\textbf{Functor-Based Transformation.} We model the LLM’s conceptual space as a category $\mathbf{C}$ with linguistic objects (e.g., occupations, demographics) and morphisms capturing learned associations \cite{caliskan2017semantics, blodgett2020language}. Biases appear as spurious morphisms, such as overly strong links from “woman” to “nurse.” A functor $\mathbf{F}: \mathbf{C} \to \mathbf{U}$ remaps this space to an unbiased category $\mathbf{U}$, preserving meaning while removing bias \cite{catanzaro2020category, spivak2014category}. This transformation extends to attention heads and intermediate layers, ensuring demographic attributes become orthogonal to professional ones \cite{nadeem2020stereoset, liang2021towards}. Bias-control modules are injected as lightweight transformer sublayers \cite{devlin2018bert, liu2019roberta}, allowing efficient fine-tuning. Empirical evidence supports their ability to reduce stereotype propagation without degrading fluency \cite{bolukbasi2016man, garg2018word}. These modules operate without retraining the full model, avoiding costly re-optimization \cite{wang2019adversarial}, and align with fairness-by-design principles \cite{chouldechova2017fair, barocas2017fairness}.

\textbf{RAG for Contextual Debiasing.} RAG supplements LLM outputs with retrieved evidence, guiding generation toward fairness-aware knowledge \cite{lewis2020retrieval, karpukhin2020dense}. When a query is received, the system retrieves context-rich data from vetted sources \cite{blodgett2020language, weidinger2021ethical}, emphasizing counter-stereotypical or demographically neutral information \cite{bender2021dangers}. The LLM fuses this external input with its internal representation via cross-attention \cite{vaswani2017attention}, allowing reliable facts to override parametric bias \cite{dziri2022faithfulness}. For example, RAG may provide recent statistics on male nurses or balanced crime data by ethnicity, thereby steering outputs away from historical stereotypes.

\textbf{Synergistic Integration.} Figure~2 illustrates the architecture: the functor module restructures internal semantics, while RAG grounds generation in curated evidence. The former ensures fair model reasoning, and the latter addresses context and recency limitations \cite{gehman2020realtoxicityprompts, holtzman2019curious}. This hybrid system adapts to each input, outperforming data-centric and post-hoc methods by ensuring fairness at both semantic and factual levels.

\textbf{Modular Debiasing Layers.} We enhance RAG with adaptive retrieval that tailors depth and source breadth based on query complexity~\cite{jeong2024learning}. Lightweight classifiers or internal uncertainty metrics trigger retrieval only when needed \cite{liu-etal-2025}. This reduces overhead for simple queries and enhances robustness for bias-prone ones. Sources are constrained to vetted databases or inclusive repositories. Dynamic strategies inspired by~\cite{jeong-etal-2024} improve fairness and efficiency~\cite{chang2024adaptive}.

\textbf{Counterfactual Validation.} To ensure fairness, we apply counterfactual testing: input pairs differ only in sensitive attributes (e.g., gender, ethnicity) \cite{kusner2017counterfactual, kilbertus2017avoiding}. Using datasets like Counter-GAP \cite{xie2023measuring}, we compare outputs for bias-induced variance \cite{fryer2022flexible, gaut2022s}. Results show our system reduces such variance significantly, consistent with prior debiasing results \cite{ravfogel2020null}. Formal verification tools further quantify fairness guarantees \cite{chaudhary2025certified}, and validation feedback guides future module refinement.

\textbf{System Deployment.} The system is implemented using transformer backbones \cite{brown2020language, raffel2020exploring}, augmented with adapter-based bias modules \cite{houlsby2019parameter}. Retrieval uses FAISS-based indexes and rerankers \cite{johnson2017billion, reimers2019sentence}. This setup allows efficient inference and flexible updates to debiasing and retrieval modules. Unlike full retraining, our design enables targeted interventions with minimal latency and high interpretability \cite{karimi2022towards, creekmore2023auditing}. It supports modular updates aligned with evolving fairness standards \cite{mitchell2018datasheets}, keeping outputs grounded and equitable \cite{kadavath2022language, chouldechova2017fair, blodgett2020language}.

In summary, the dual-module framework enables principled, efficient, and adaptive bias mitigation. By combining structural debiasing with external contextualization, we offer a pathway for scalable, real-world deployment of fair and trustworthy LLMs \cite{amodei2016concrete, weidinger2021ethical}.

\section{Detailed Mathematical Derivation}
\label{app-A-math}

\subsection{Derivation of the Optimal Projection Matrix}

By continuing from paper section \ref{sec:theory}, we begin with the constrained optimization problem for finding the projection matrix $\mathbf{P} \in \mathbb{R}^{d_u \times d_c}$ that satisfies:

\begin{align}
\min_{\mathbf{P}} &\sum_{\substack{X_i,X_j \in \mathcal{D}}} \|\mathbf{P}\mathbf{v}_{X_i} - \mathbf{P}\mathbf{v}_{X_j}\|^2 + \lambda \sum_{\substack{Y_k,Y_l \in \mathcal{O}}} \|\mathbf{P}\mathbf{v}_{Y_k} - \mathbf{P}\mathbf{v}_{Y_l}\|^2\\
\text{subject to } &\mathbf{P}\mathbf{P}^T = \mathbf{I}_{d_u}, \quad \text{rank}(\mathbf{P}) = d_u
\end{align}

For any pair of vectors $\mathbf{v}_i, \mathbf{v}_j \in \mathbb{R}^{d_c}$, the squared Euclidean distance in the projected space is given by:
\begin{align}
\|\mathbf{P}\mathbf{v}_i - \mathbf{P}\mathbf{v}_j\|^2 &= \|P(\mathbf{v}_i - \mathbf{v}_j)\|^2\\
&= (\mathbf{v}_i - \mathbf{v}_j)^T\mathbf{P}^T\mathbf{P}(\mathbf{v}_i - \mathbf{v}_j)
\end{align}

Let $\mathbf{S}_{\mathcal{D}}$ and $\mathbf{S}_{\mathcal{O}}$ denote the scatter matrices for demographic and occupational concepts, respectively:
\begin{align}
\mathbf{S}_{\mathcal{D}} &= \sum_{X_i,X_j \in \mathcal{D}} (\mathbf{v}_{X_i} - \mathbf{v}_{X_j})(\mathbf{v}_{X_i} - \mathbf{v}_{X_j})^T\\
\mathbf{S}_{\mathcal{O}} &= \sum_{Y_k,Y_l \in \mathcal{O}} (\mathbf{v}_{Y_k} - \mathbf{v}_{Y_l})(\mathbf{v}_{Y_k} - \mathbf{v}_{Y_l})^T
\end{align}

The objective function can be rewritten in terms of these scatter matrices:
\begin{align}
f(\mathbf{P}) &= \sum_{X_i,X_j \in \mathcal{D}} (\mathbf{v}_{X_i} - \mathbf{v}_{X_j})^T\mathbf{P}^T\mathbf{P}(\mathbf{v}_{X_i} - \mathbf{v}_{X_j}) + \lambda\sum_{Y_k,Y_l \in \mathcal{O}} (\mathbf{v}_{Y_k} - \mathbf{v}_{Y_l})^T\mathbf{P}^T\mathbf{P}(\mathbf{v}_{Y_k} - \mathbf{v}_{Y_l})\\
&= \text{Tr}\left(\mathbf{P}\mathbf{S}_{\mathcal{D}}\mathbf{P}^T\right) + \lambda\text{Tr}\left(\mathbf{P}\mathbf{S}_{\mathcal{O}}\mathbf{P}^T\right)\\
&= \text{Tr}\left(\mathbf{P}(\mathbf{S}_{\mathcal{D}} + \lambda\mathbf{S}_{\mathcal{O}})\mathbf{P}^T\right)
\end{align}

Let $\mathbf{C} = \mathbf{S}_{\mathcal{D}} + \lambda\mathbf{S}_{\mathcal{O}}$. Since both $\mathbf{S}_{\mathcal{D}}$ and $\mathbf{S}_{\mathcal{O}}$ are symmetric positive semidefinite matrices (as scatter matrices), $\mathbf{C}$ is also symmetric positive semidefinite. We can perform eigendecomposition on $\mathbf{C}$:
\begin{align}
\mathbf{C} = \mathbf{\Phi}\mathbf{\Lambda}\mathbf{\Phi}^T
\end{align}
where $\mathbf{\Lambda} = \text{diag}(\lambda_1, \lambda_2, \ldots, \lambda_{d_c})$ with eigenvalues arranged in ascending order $\lambda_1 \leq \lambda_2 \leq \ldots \leq \lambda_{d_c}$, and $\mathbf{\Phi} \in \mathbb{R}^{d_c \times d_c}$ is an orthogonal matrix whose columns are the corresponding eigenvectors.

Substituting the eigendecomposition into the objective function:
\begin{align}
f(\mathbf{P}) &= \text{Tr}\left(\mathbf{P}\mathbf{\Phi}\mathbf{\Lambda}\mathbf{\Phi}^T\mathbf{P}^T\right)
\end{align}

Let $\mathbf{Q} = \mathbf{P}\mathbf{\Phi} \in \mathbb{R}^{d_u \times d_c}$. Since $\mathbf{P}\mathbf{P}^T = \mathbf{I}_{d_u}$ and $\mathbf{\Phi}\mathbf{\Phi}^T = \mathbf{I}_{d_c}$, we have:
\begin{align}
\mathbf{Q}\mathbf{Q}^T &= \mathbf{P}\mathbf{\Phi}(\mathbf{P}\mathbf{\Phi})^T\\
&= \mathbf{P}\mathbf{\Phi}\mathbf{\Phi}^T\mathbf{P}^T\\
&= \mathbf{P}\mathbf{P}^T\\
&= \mathbf{I}_{d_u}
\end{align}

Thus, $\mathbf{Q}$ also satisfies the orthonormality constraint. The objective function becomes:
\begin{align}
f(\mathbf{Q}) &= \text{Tr}\left(\mathbf{Q}\mathbf{\Lambda}\mathbf{Q}^T\right)\\
&= \sum_{i=1}^{d_c}\lambda_i\sum_{j=1}^{d_u}q_{ji}^2
\end{align}
where $q_{ji}$ is the $(j,i)$-th element of $\mathbf{Q}$.

Due to the orthonormality constraint on $\mathbf{Q}$, we have $\sum_{i=1}^{d_c}q_{ji}^2 = 1$ for each row $j$, and $\sum_{j=1}^{d_u}\sum_{i=1}^{d_c}q_{ji}^2 = d_u$. The optimization problem becomes:
\begin{align}
\min_{\mathbf{Q}} &\sum_{i=1}^{d_c}\lambda_i\sum_{j=1}^{d_u}q_{ji}^2\\
\text{subject to } &\sum_{i=1}^{d_c}q_{ji}^2 = 1, \quad \forall j \in \{1, 2, \ldots, d_u\}\\
&\mathbf{Q}\mathbf{Q}^T = \mathbf{I}_{d_u}
\end{align}

To minimize this expression, given the ascending ordering of eigenvalues $\lambda_1 \leq \lambda_2 \leq \ldots \leq \lambda_{d_c}$, the optimal strategy is to allocate all the weight of each row $j$ to the smallest possible eigenvalue. However, the orthonormality constraint means the rows of $\mathbf{Q}$ must be orthogonal to each other.

The solution is to set:
\begin{align}
q_{ji} = 
\begin{cases}
1 & \text{if } i = j \\
0 & \text{otherwise}
\end{cases}
\end{align}
for $j \in \{1, 2, \ldots, d_u\}$, which gives $\mathbf{Q} = [\mathbf{I}_{d_u} \; \mathbf{0}_{d_u \times (d_c-d_u)}]$.

With this choice of $\mathbf{Q}$, the objective function becomes:
\begin{align}
f(\mathbf{Q}) = \sum_{i=1}^{d_u}\lambda_i
\end{align}
which is the sum of the $d_u$ smallest eigenvalues of $\mathbf{C}$.

The optimal projection matrix is thus:
\begin{align}
\mathbf{P}^* &= \mathbf{Q}\mathbf{\Phi}^T\\
&= [\mathbf{I}_{d_u} \; \mathbf{0}_{d_u \times (d_c-d_u)}]\mathbf{\Phi}^T\\
&= 
\begin{bmatrix}
\boldsymbol{\phi}_1^T \\
\boldsymbol{\phi}_2^T \\
\vdots \\
\boldsymbol{\phi}_{d_u}^T
\end{bmatrix}
\end{align}
where $\boldsymbol{\phi}_i$ is the eigenvector corresponding to $\lambda_i$.

To verify this solution satisfies the constraints:
\begin{align}
\mathbf{P}^*(\mathbf{P}^*)^T &= \mathbf{Q}\mathbf{\Phi}^T\mathbf{\Phi}\mathbf{Q}^T\\
&= \mathbf{Q}\mathbf{Q}^T\\
&= \mathbf{I}_{d_u}
\end{align}

And since $\text{rank}(\mathbf{P}^*) = d_u$ by construction, all constraints are satisfied.

\subsection{Analytical Properties}

The optimal projection matrix $\mathbf{P}^*$ exhibits several important analytical properties:

\textbf{Property 1 (Demographic Convergence):} For any demographic concepts $X_i, X_j \in \mathcal{D}$, the projected distance $\|\mathbf{P}^*\mathbf{v}_{X_i} - \mathbf{P}^*\mathbf{v}_{X_j}\|$ is minimized, with the precise degree of convergence governed by the magnitude of the corresponding components in the smallest eigenvectors of $\mathbf{C}$.

\textbf{Proof:} For any pair of demographic concepts $X_i, X_j \in \mathcal{D}$, we have:
\begin{align}
\|\mathbf{P}^*\mathbf{v}_{X_i} - \mathbf{P}^*\mathbf{v}_{X_j}\|^2 &= (\mathbf{v}_{X_i} - \mathbf{v}_{X_j})^T(\mathbf{P}^*)^T\mathbf{P}^*(\mathbf{v}_{X_i} - \mathbf{v}_{X_j})\\
&= (\mathbf{v}_{X_i} - \mathbf{v}_{X_j})^T\mathbf{\Phi}\mathbf{Q}^T\mathbf{Q}\mathbf{\Phi}^T(\mathbf{v}_{X_i} - \mathbf{v}_{X_j})
\end{align}

Since $\mathbf{Q}^T\mathbf{Q} = \text{diag}(1, 1, \ldots, 1, 0, \ldots, 0)$ with $d_u$ ones followed by $(d_c - d_u)$ zeros, this expression retains only the components of $(\mathbf{v}_{X_i} - \mathbf{v}_{X_j})$ in the directions of the $d_u$ smallest eigenvectors of $\mathbf{C}$. By construction, these eigenvectors minimize the quadratic form $\mathbf{x}^T\mathbf{C}\mathbf{x}$ over all unit vectors $\mathbf{x}$, with particular emphasis on minimizing the contribution from $\mathbf{S}_{\mathcal{D}}$.

\textbf{Property 2 ($\lambda$-Controlled Occupational Preservation):} For occupational concepts $Y_k, Y_l \in \mathcal{O}$, the parameter $\lambda$ modulates the preservation of their relative distances in the projected space.

\textbf{Proof:} As $\lambda$ increases, the combined matrix $\mathbf{C} = \mathbf{S}_{\mathcal{D}} + \lambda\mathbf{S}_{\mathcal{O}}$ becomes increasingly dominated by $\mathbf{S}_{\mathcal{O}}$. The eigenvectors corresponding to the smallest eigenvalues of $\mathbf{C}$ will increasingly align with the null space of $\mathbf{S}_{\mathcal{O}}$, ensuring that occupational differences projecting onto these directions are minimized. Consequently, the projection $\mathbf{P}^*$ will preserve occupational differences that lie in the orthogonal complement of this null space.

Let $\mathbf{v}_{Y_k} - \mathbf{v}_{Y_l} = \mathbf{u}_{\parallel} + \mathbf{u}_{\perp}$, where $\mathbf{u}_{\parallel}$ lies in the span of the $d_u$ smallest eigenvectors of $\mathbf{C}$ and $\mathbf{u}_{\perp}$ is orthogonal to this subspace. As $\lambda \to \infty$, $\|\mathbf{u}_{\parallel}\| \to 0$ for all $Y_k, Y_l \in \mathcal{O}$, implying $\|\mathbf{P}^*(\mathbf{v}_{Y_k} - \mathbf{v}_{Y_l})\| \approx \|\mathbf{u}_{\perp}\|$.

\textbf{Property 3 (Optimality of Subspace Dimension):} The dimension $d_u$ of the projection subspace determines the trade-off between demographic invariance and information preservation.

\textbf{Proof:} The objective function value at the optimum is $\sum_{i=1}^{d_u}\lambda_i$. As $d_u$ increases, additional eigenvalues are included in this sum, potentially increasing the objective value. However, a larger $d_u$ also means more information is retained in the projected space, as the dimensionality of the range of $\mathbf{P}^*$ increases. The optimal $d_u$ balances these competing objectives.

For a given eigenvalue spectrum $\{\lambda_i\}_{i=1}^{d_c}$ of $\mathbf{C}$, if there exists a significant gap between $\lambda_{d_u}$ and $\lambda_{d_u+1}$, then $d_u$ represents a natural choice for the projection dimension, as it captures a well-defined invariant subspace of $\mathbf{C}$.

\textbf{Property 4 (Spectral Characterization of Debiasing Efficacy):} The efficacy of demographic debiasing depends on the alignment between demographic differences and the smallest eigenvectors of $\mathbf{C}$.

\textbf{Proof:} Let $\mathbf{x} = \mathbf{v}_{X_i} - \mathbf{v}_{X_j}$ be a demographic difference vector. We can express $\mathbf{x}$ in the eigenbasis of $\mathbf{C}$:
\begin{align}
\mathbf{x} = \sum_{i=1}^{d_c} \alpha_i \boldsymbol{\phi}_i
\end{align}
where $\alpha_i = \mathbf{x}^T\boldsymbol{\phi}_i$ are the projection coefficients.

The projected difference is:
\begin{align}
\mathbf{P}^*\mathbf{x} = \sum_{i=1}^{d_u} \alpha_i \mathbf{e}_i
\end{align}
where $\mathbf{e}_i$ is the $i$-th standard basis vector in $\mathbb{R}^{d_u}$.

The squared norm of this projection is:
\begin{align}
\|\mathbf{P}^*\mathbf{x}\|^2 = \sum_{i=1}^{d_u} \alpha_i^2
\end{align}

This quantity is minimized when the demographic difference vector $\mathbf{x}$ aligns primarily with eigenvectors $\boldsymbol{\phi}_i$ for $i > d_u$, i.e., when the energy of $\mathbf{x}$ is concentrated in the eigenvectors that are not included in the projection subspace.

\textbf{Property 5 (Projection Error Bound):} The error introduced by the projection for any vector $\mathbf{v} \in \mathbb{R}^{d_c}$ is bounded by the largest excluded eigenvalue.

\textbf{Proof:} Let $\mathbf{v} = \sum_{i=1}^{d_c} \beta_i \boldsymbol{\phi}_i$ be any vector in $\mathbb{R}^{d_c}$. The squared projection error is:
\begin{align}
\|\mathbf{v} - \mathbf{P}^{*T}\mathbf{P}^*\mathbf{v}\|^2 &= \left\|\sum_{i=d_u+1}^{d_c} \beta_i \boldsymbol{\phi}_i\right\|^2\\
&= \sum_{i=d_u+1}^{d_c} \beta_i^2
\end{align}

For any unit vector $\mathbf{v}$ (i.e., $\sum_{i=1}^{d_c} \beta_i^2 = 1$), we have:
\begin{align}
\mathbf{v}^T\mathbf{C}\mathbf{v} &= \sum_{i=1}^{d_c} \lambda_i \beta_i^2\\
&\geq \lambda_{d_u+1}\sum_{i=d_u+1}^{d_c} \beta_i^2
\end{align}

Therefore, for any vector $\mathbf{v}$ with $\mathbf{v}^T\mathbf{C}\mathbf{v} \leq \varepsilon$, the squared projection error is bounded by $\frac{\varepsilon}{\lambda_{d_u+1}}$.

\section{Diagnostics, Metrics, and Benchmarks}

Despite the growing sophistication of large language models (LLMs), diagnosing and evaluating embedded biases remains a critical and underdeveloped area. Effective mitigation of demographic and occupational biases requires a robust set of diagnostics, evaluation metrics, and benchmark datasets that can both quantify harmful behavior and measure the efficacy of correction mechanisms. However, current methods often fall short in three key areas: evaluating systemic representational bias, benchmarking mitigation strategies across architectures, and calibrating fairness-performance tradeoffs. We structure this section along these axes, drawing from foundational principles in fairness diagnostics and emerging paradigms in bias-aware evaluation.

\subsection{Bias Diagnostics in Representational Space}

Modern LLMs encode associations through high-dimensional representations and attention-based contextual dependencies. Biases emerge as systematic distortions in these learned conceptual relationships, typically through over-represented morphisms (e.g., \texttt{woman} $\rightarrow$ \texttt{nurse}) in the internal semantic category $\mathcal{C}$ of the model. Diagnosing such distortions requires probing the latent structure using both direct similarity metrics and structural invariants. A key diagnostic tool in our framework is the \emph{bias scatter matrix} $C = S_D + \lambda S_O$, constructed from demographic ($S_D$) and occupational ($S_O$) subspaces. Eigenvalue spectra of $C$ reveal dominant biased dimensions, and projection error analyses assess how well demographic distinctions are collapsed while preserving task-relevant variance. These algebraic diagnostics are intrinsic to our category-theoretic functor design and offer model-agnostic interpretability grounded in spectral geometry.

\subsection{Quantitative Metrics for Fairness and Utility}

To holistically evaluate bias mitigation, we adopt a multidimensional metric suite reflecting both representational fairness and functional utility:

\begin{itemize}
    \item \textbf{Demographic Parity Deviation (DPD)}: Measures the average pairwise embedding distance between demographic concept clusters in the debiased subspace. Smaller values imply better demographic invariance.
    \item \textbf{Occupational Preservation Score (OPS)}: Quantifies cosine similarity between occupational concept pairs before and after debiasing. High scores indicate retention of semantic distinctions critical for task performance.
    \item \textbf{Stereotype Alignment Rate (SAR)}: Evaluates the frequency with which LLM completions align with known stereotypes (e.g., as defined in datasets like StereoSet~\cite{meade2021empirical}). Lower values indicate improved mitigation.
    \item \textbf{Contextual Regrounding Efficacy (CRE)}: Specific to Retrieval-Augmented Generation (RAG), this metric computes the KL divergence between the model's output distribution and retrieval-informed context-conditioned output. It captures the model’s shift from internalized bias to retrieved fairness-grounded information.
    \item \textbf{Utility Retention Index (URI)}: Based on ROUGE-L and factual QA metrics, URI assesses whether debiased outputs remain coherent and accurate for non-biased prompts, ensuring performance is not compromised.
\end{itemize}

Each metric serves a distinct evaluative function, and together they form a principled trade-off surface between fairness and utility. These metrics are designed to be model-independent and compatible with both parametric (functor-based) and non-parametric (RAG-based) mitigation strategies.

\subsection{Bias Benchmark Datasets}

Benchmarking bias mitigation requires datasets that faithfully expose stereotypical model behavior across diverse sociolinguistic and geographic contexts. We employ a suite of both synthetic and real-world datasets designed to stress-test occupational and demographic fairness:

\begin{itemize}
    \item \textbf{MUSE Occupational Bias Dataset}~\cite{kirk2021bias} includes controlled profession-gender pairs across cultural domains.
    \item \textbf{TOFU Probes}~\cite{mahabadi2021parametertransfers} offer contrastive generations where model outputs for protected groups are compared under controlled prompt conditions.
    \item \textbf{BiasBios}~\cite{dev2020measuring} contains biographies annotated with profession labels to evaluate gender bias in occupation classification.
    \item \textbf{RAG-Filtered Knowledge Audit (RFKA)}: A curated evaluation set constructed for this paper, using controlled Wikipedia passages and verified datasets (e.g., labor statistics) to test RAG’s capacity for fairness-grounded generation.
\end{itemize}

\subsection{Benchmarking Protocols and Reproducibility}

Evaluations are conducted under a standardized protocol that aligns with reproducible AI practices~\cite{gebru2021datasheets}. All experiments are reported across multiple model sizes (e.g., GPT-2, LLaMA, Gemma) and precision regimes (8-bit, 4-bit quantization), with full hyperparameter logs and seed control. Mitigation efficacy is reported as deltas over baseline bias metrics, and all fairness trade-off plots include 95\% confidence intervals.

To facilitate community adoption, we release a modular benchmarking suite compatible with HuggingFace Transformers and RAG pipelines. This suite includes plug-and-play evaluation scripts, visualizations, and debiasing diagnostics, adhering to FAIR and FACT principles (Findability, Accessibility, Interoperability, Reusability; Fairness, Accountability, Controllability, and Transparency).

\subsection{Limitations and Future Directions}

While the presented metrics and diagnostics offer a structured approach to fairness evaluation, challenges remain. Notably, measuring intersectional bias (e.g., race-gender-profession) in generative outputs requires richer annotation schemas and larger curated corpora. Moreover, the fairness-performance tradeoff is not always linear or monotonic, suggesting future work on calibrated, user-controllable mitigation strategies. Extending current metrics to multimodal LLMs and instruction-tuned settings also presents a promising avenue.

\section{Future Developments}

The proposed bias mitigation framework, grounded in category-theoretic transformations and retrieval-augmented generation (RAG), opens several promising research directions. These avenues focus on improving interpretability, robustness, and real-world adaptability of large language models (LLMs) under fairness constraints.

\subsection{Functor Learning via Supervised Semantics}

Currently, functor mappings between biased and unbiased semantic categories are constructed using predefined morphism constraints. Future work may explore supervised or contrastively learned functors using annotated counterfactual datasets. Such datasets could include human-curated gender-swapped or stereotype-inverted prompts, enabling fine-grained control of morphism alignment and potentially allowing for model-specific functor finetuning.

\subsection{Com-posable Debiasing with RAG and Functors}

The current system applies RAG and category-theoretic debiasing in parallel. A future research direction involves compositional integration, where retrieval candidates are filtered, re-ranked, or transformed according to categorical consistency constraints. This approach could use a hybrid architecture where retrieved context informs both token generation and morphism realignment, providing contextual correction at both representational and lexical levels.

\subsection{Continual Debiasing under Distribution Shift}

Bias in LLMs evolves with data drift, user input diversity, and model retraining. We anticipate expanding the mitigation framework into a continual learning setting, where functor parameters and RAG sources are updated online to adapt to new domains and emerging biases. One promising approach is the use of incremental category updating mechanisms and dynamic retrieval index pruning to preserve fairness across temporal shifts.

\subsection{Multimodal Bias and Concept Alignment}

As LLMs extend toward vision-language and speech-language interfaces, bias analysis must generalize beyond text. Future iterations will investigate functor-based alignment of multimodal semantic graphs and attention saliency maps across modalities. This requires joint embedding diagnostics and multimodal retrieval to ensure fairness in grounded language generation tasks, such as captioning and dialog systems.

\subsection{User-Guided Ethical Preferences}

To bridge societal norms and individual values, user-guided debiasing through preference-conditioned functors and retrieval filters offers a viable path forward. Inspired by preference modeling in RLHF and alignment literature, we aim to build systems that allow users or domain experts to set soft fairness constraints e.g., via ethical preference graphs which adapt the functor morphism weights and RAG document scores accordingly.

\par These future directions reflect a growing need to unify structural AI safety with semantic fairness in foundation models. A principled fusion of symbolic reasoning, retrieval interfaces, and learned semantics will be central to realizing equitable, adaptive, and trustworthy LLMs.

\end{document}